\documentclass[reprint,amsmath,amssymb,aps]{revtex4-1}

\usepackage{booktabs}
\usepackage{nicefrac}
\usepackage{hyperref}
\hypersetup{%
hidelinks,
}
\usepackage{url}

\newcommand{\B}[1]{{\boldsymbol{#1}}}
\newcommand{\mac}[1]{{\mathcal #1}}
\newcommand{\mab}[1]{{\mathbb #1}}

\usepackage{stackengine}
\def\delequal{\mathrel{\ensurestackMath{\stackon[1pt]{=}{\scriptstyle\Delta}}}}

\usepackage{mathtools}
\usepackage{amsthm}

\usepackage[labelformat=simple]{subcaption}
\usepackage{graphicx}  
\usepackage{overpic}
\newcommand{\subfiglabelsize}{}

\theoremstyle{plain}
\newtheorem{theorem}{Theorem}[section]

\theoremstyle{definition}

\theoremstyle{remark}

\begin{document}

\title{Ratio Divergence Learning Using Target Energy in Restricted Boltzmann Machines: \\
Beyond Kullback--Leibler Divergence Learning}

\author{Yuichi Ishida$^{1,}$}
\email{ishida-yuichi@fujitsu.com}
\author{Yuma Ichikawa$^{1, 2,}$}
\email{ichikawa.yuma@fujitsu.com}
\author{Aki Dote$^{1}$}
\email{dote.aki@fujitsu.com}
\author{Toshiyuki Miyazawa$^{1}$}
\email{miyazawa.toshi@fujitsu.com}
\author{Koji Hukushima$^{1, 2}$}
\email{k-hukushima@g.ecc.u-tokyo.ac.jp}
\affiliation{$^{1}$Fujitsu, Kawasaki, Japan}
\affiliation{$^{2}$Graduate School of Arts and Sciences, University of Tokyo, Tokyo, Japan}

\begin{abstract}
    We propose ratio divergence (RD) learning for discrete energy-based models, a method that utilizes both training data and a tractable target energy function.
    We apply RD learning to restricted Boltzmann machines (RBMs), which are a minimal model that satisfies the universal approximation theorem for discrete distributions.
    RD learning combines the strength of both forward and reverse Kullback--Leibler divergence (KLD) learning, effectively addressing the ``notorious'' issues of underfitting with the forward KLD and mode-collapse with the reverse KLD.
    Since the summation of forward and reverse KLD seems to be sufficient to combine the strength of both approaches, we include this learning method as a direct baseline in numerical experiments to evaluate its effectiveness.
    Numerical experiments demonstrate that RD learning significantly outperforms other learning methods in terms of energy function fitting, mode-covering, and learning stability across various discrete energy-based models.
    Moreover, the performance gaps between RD learning and the other learning methods become more pronounced as the dimensions of target models increase.
\end{abstract}

\maketitle

\section{Introduction}\label{sec:introduction}
Discrete high-dimensional distributions, where the energy function is tractable but the normalization factor is not, appear in various fields, including condensed matter physics \citep{binder1993monte, baumgartner2012monte}, Bayesian inference \citep{martin2011mcmcpack, gelman1995bayesian, foreman2013emcee}, and optimization \citep{dabiri2020replica}.
For these applications, it is crucial to understand the statistical properties of these distributions and to draw samples from them efficiently.
However, these tasks often present considerable computational issues, even when the energy function is tractable.

Therefore, approximating target distributions using generative models, which can generate samples with low computational cost, has gained attention. 
This approach has been successfully applied in various \textit{downstream tasks}, such as low-dimensional compression \citep{gu2022thermodynamics}, discovery of critical clusters  \citep{chen2018collective, ribeiro2018reweighted}, enhancement of Markov Chain Monte Carlo (MCMC) simulations \citep{liu2017self, huang2017accelerated, xu2017self}, and molecular dynamics  \citep{wu2020stochastic, wirnsberger2022normalizing, stimper2022resampling}.
Learning methods for these generative models can be categorized into two types based on the Kullback--Leibler (KL) divergence: forward KLD learning, which employs only samples drawn from the target distribution, and reverse KLD learning, which uses the target energy function.
However, \citet{ciarella2023machine} identified limitations in both the learning methods when applied to complex distributions with exponential peaks.
While the forward KLD learning results in ``underfitting'' of the target energy function, particularly in the presence of multiple peaks, the reverse KLD learning results in ``mode collapse'', even in simple probability distributions.

To address these issues, we propose a learning method using \textit{ratio divergence (RD)}, a symmetrized divergence that naturally incorporates the product of the forward and reverse KLDs while eliminating the intractable normalization factor of the target energy.
Although a linear combination of the forward and reverse KLDs might seem sufficient to address the issues, it is difficult to determine coefficients.
Additionally, minimizing RD has the advantage of increasing the acceptance probability of the Metropolis-Hastings algorithm in MCMC simulations, which is one of the downstream tasks.
We then apply RD learning to the restricted Boltzmann machine (RBM) \citep{smolensky1986information, hinton2002training}, a minimal model satisfying the universal approximation theorem \citep{le2008representational} and employed for numerous downstream tasks \citep{huang2017accelerated, hjelm2014restricted, melko2019restricted}.

Numerical experiments demonstrate that RD learning more effectively approximates target distributions than the forward and reverse KLD learning methods in various discrete models, such as the ferromagnetic Ising model, the Sherrington--Kirkpatrick (SK) model, the maximum independent set (MIS) problem, and the maximum cut problem (MCP) on various graphs. 
Particularly for the MIS, which is known for its exponential number of local optima and is noted by \citet{ciarella2023machine} that approximating the distribution is difficult using autoregressive generative models \citep{khajenezhad2020masked, uria2016neural}, RD learning achieves superior performance in both energy fitting and mode covering abilities compared to the forward and reverse KLD learning. 
These results are also consistent with the MCPs on diverse graphs.
Furthermore, our experiments show that RD learning is consistently effective for high-dimensional distributions, where its performance gap for the forward and reverse KLD learning becomes more significant as the dimension of target energy increases.

This paper is organized as follows.
Sec.~\ref{sec:background} introduces the structure of RBM and its learning methods based on forward and reverse KLDs.
Sec.~\ref{sec:method} discusses the practical issues with these learning methods and proposes a learning method based on RD, emphasizing its features.
Sec.~\ref{sec:related-work} reviews other learning methods related to our approach.
Sec.~\ref{sec:experiments} presents numerical experiments, applying them to several example models to compare the learning methods based on KLDs and RD.
Finally, the conclusion is provided in Sec.~\ref{sec:conclusion}. 

\section{Background}\label{sec:background}
\subsection{Problem setting}\label{sec:problem-setting}
We begin by explaining the problem settings of this study.
Let $\hat{P} (\B{x}) = \exp(-\hat{E}(\B{x}))/\hat{Z}$ represent the target discrete distribution over a discrete state space $\mac{X}$, where $\hat{Z} \delequal \sum_{\B{x}\in\mac{X}} \exp(-\hat{E}(\B{x}))$ is the normalization factor.
While computing the energy $\hat{E}(\B{x})$ for a given $\B{x}\in \mac{X}$ is relatively inexpensive, calculating the normalization factor $\hat{Z}$ and the expected value with respect to $\hat{P}(\B{x})$ can be computationally challenging. 
In the following, let $\mathcal{D} = \{\B{x}_\mu\}_{\mu=1}^{M}, ~\B{x}_\mu \sim \hat{P}$ denote training dataset with $M$ being the number of the data. 
$\hat{P}_\mathcal{D}$ denotes the empirical distribution.
Under these settings, this study aims to approximate this target distribution using a machine-learning model.

\subsection{Restricted Boltzmann machines}\label{subsec:rbm}
A Boltzmann machine is an energy-based model consisting of two layers of binary variables: $\B{x} \in \{0, 1\}^{N_{x}}$ and $\B{h} \in \{0, 1\}^{N_{h}}$, which are referred to as visible and hidden layers, respectively.
The probability distribution over these variables is given by
\begin{subequations}
    \label{eq:rbm-boltzmann-measure}
    \begin{gather}
        P(\B{x}, \B{h}; \B{\theta}) = \frac{1}{Z(\B{\theta})} e^{- E(\B{x}, \B{h}; \B{\theta})},\\
        Z(\B{\theta}) \delequal \sum_{\B{x}, \B{h}} e^{- E(\B{x}, \B{h}; \B{\theta})},
    \end{gather}
\end{subequations}
where $E(\B{x}, \B{h}; \B{\theta})$ represents the energy function of the model.
To accelerate both training and inference, an energy function that includes interactions only between different layers, with no interactions within the same layer, is known as a Restricted Boltzmann Machine \citep{smolensky1986information, hinton2002training}.
Specifically, The energy function is defined as
\begin{equation}
        E(\B{x}, \B{h}; \B{\theta}) = - \B{b}^\top \B{x} - \B{c}^\top \B{h} - \B{x}^\top W \B{h},
\end{equation}
where $W \in \mab{R}^{N_{x} \times N_{h}}$, $\B{b} \in \mab{R}^{N_{x}}$, and $\B{c} \in \mab{R}^{N_{h}}$ are learning parameters collectively denoted as $\B{\theta}$.
Under this restriction, the marginal distribution $P(\B{x}; \B{\theta})$ is expressed as: 
\begin{subequations}
    \label{eq:free-energy-visibe}
    \begin{gather}
        P(\B{x}; \B{\theta}) = \frac{1}{Z(\B{\theta})}e^{-F(\B{x}; \B{\theta})},\\
        F(\B{x};\B{\theta}) = -\B{b}^{\top} \B{x} - \sum_{m=1}^{N_{h}} \ln \left(1 + e^{c_{m}+ \sum_{i=1}^{N} x_{i} W_{im}}\right).
    \end{gather}
\end{subequations}
The learning method of RBMs can be categorized into two methods.

\paragraph*{(I) Forward KLD learning}
One learning approach involves training RBMs by minimizing the forward KLD defined by
\begin{equation}
    D_{\mathrm{KL}}[\hat{P}_{\mac{D}}(\B{x}) \| P(\B{x}; \B{\theta})]
    = \mab{E}_{\mac{D}} \left[ \ln \frac{\hat{P}_{\mac{D}}(\B{x})}{P(\B{x};\B{\theta})} \right],
    \label{eq:fowardKL}
\end{equation}
where $\mab{E}_{\mac{D}}[f(\B{x})] = M^{-1} \sum_{\mu=1}^{M} f(\B{x})$ denotes the average over the dataset.
The gradients of the forward KLD with respect to the parameters are given by 
\begin{multline}
    \label{eq:rbm-gradient}
    \frac{\partial D_{\mathrm{KL}}[\hat{P}_{\mac{D}}(\B{x}) \| P(\B{x}; \B{\theta})]}{\partial \B{\theta}} \\
    = \mathbb{E}_{p(\B{x}, \B{h}; \B{\theta})}\left[ \frac{\partial E(\B{x}, \B{h}; \B{\theta})}{\partial \B{\theta}} \right] 
    -\mab{E}_{\mac{D}} \mab{E}_{p(\B{h}|\B{x})} \left[ \frac{\partial E(\B{x}, \B{h}; \B{\theta})}{\partial \B{\theta}}  \right],
\end{multline}
where $\mab{E}_{p(\B{h}|\B{x})}[f(\B{x}, \B{h})] =  \sum_{\B{h}} f(\B{x}, \B{h})p(\B{h}|\B{x})$
denotes the average over the hidden variables for given $\B{x}$ and $\mab{E}_{p(\B{x}, \B{h};\B{\theta})}[f(\B{x}, \B{h})]$ denotes the average over the Boltzmann measure.
Minimizing this divergence is computationally challenging due to the intractable summation of all variables on the right-hand side of Eq.~\eqref{eq:rbm-gradient}.
Therefore, these expectation values are approximated by the \textit{self-samples} drawn from the Boltzmann measure in Eq.~\eqref{eq:rbm-boltzmann-measure}.
One of the most widely used approximation methods is the persistent contrastive divergence (PCD) method~\citep{hinton2002training,Tieleman_2008_Proc.25thInt.Conf.Mach.Learn.}, in which the right-hand side of the gradient is estimated after a few $k$ steps of the block Gibbs sampling (BGS).
In practice, the step is typically with $k=1$.
The final state of the BGS under the current parameters is then used as the initial state for the next BGS after the parameters are updated. 

\paragraph*{(II) Reverse KLD learning}
The other approach is minimizing the reverse KLD defined by
\begin{equation}
    D_{\mathrm{KL}}[P(\B{x}; \B{\theta}) \| \hat{P}_{\mac{D}}(\B{x})]
    = \mab{E}_{P(\B{x};\B{\theta})} \left[ \ln \frac{P(\B{x};\B{\theta})}{\hat{P}_{\mac{D}}(\B{x})} \right].
    \label{eq:reverseKL}
\end{equation}
\citet{puente2020convolutional} proved that the gradients of the reverse KLD are equivalent to those of the following loss function:
\begin{subequations}
    \begin{gather}
        \mab{E}_{P(\B{x};\B{\theta})} \left[ \frac{1}{2} \left( \hat{E}(\B{x}) - F(\B{x}) - C(\B{\theta}) \right)^2 \right], \\
        C(\B{\theta}) = \mab{E}_{P(\B{x};\B{\theta})} \left[ \left( \hat{E}(\B{x}) - F(\B{x}) \right)^2 \right].
    \end{gather}
\end{subequations}
To compute this loss function and its gradients, these expectation values over the Boltzmann measure are approximated by the self-samples drawn from the Boltzmann measure, just as when calculating the gradients of the forward KLD.
Note that this method is only applicable when the target energy function is explicitly provided.

\section{Method}\label{sec:method}
In this section, we identify two practical issues associated with forward and reverse KLD learning. 
To address these challenges, we propose a learning method, \textit{ratio divergence (RD) learning}, which we define and discuss below.

\subsection{Practical issues of forward and reverse KLD learning}\label{subsec:issues}

\paragraph*{Underfitting in forward KLD learning}
Forward KLD learning relies solely on the training dataset $\mathcal{D}$ and does not utilize the target energy function $\hat{E}(\B{x})$ even when $\hat{E}(\B{x})$ is provided. 
As a result, the regression performance for the target energy is not improved with training models.
Indeed, \citet{ciarella2023machine} demonstrated that while forward KLD learning can cover many peaks of the target probability distribution $\hat{P}(\B{x})$ (i.e., mode-covering), it fails to capture the detailed structure of these peaks, which adversely affects downstream tasks.

\paragraph*{Mode collapse in reverse KLD learning}
Reverse KLD learning relies solely on the target energy function $\hat{E}(\B{x})$ and does not leverage the training dataset $\hat{\mathcal{D}}$.
For this reason, once a high-probability sample is found, it becomes difficult to generate samples far from it.
As a result, reverse KLD learning can approximate only one or a few peaks of the target probability distribution $\hat{P}(\B{x})$.
This phenomenon is a frequently observed issue in reverse KLD learning, commonly referred to as a mode-collapse by \citet{ciarella2023machine}.

\subsection{Ratio divergence learning}\label{subsec:ratio-divergence}
To address these two issues, we propose a learning method that minimizes a symmetrized divergence, termed \textit{ratio divergence}.
RD effectively integrates both the forward and reverse KLDs as follows:
\begin{equation}
\label{eq:ratio-divergence}
    \mac{L}(\hat{P}, P;\B{\theta}) = \sum_{\B{x}^{\prime}, \B{x} \in \mac{X}}  \hat{P}(\B{x}^{\prime}) P(\B{x}; \B{\theta}) \left(\log \frac{\hat{P}(\B{x}^{\prime}) P(\B{x};\B{\theta})}{P(\B{x}^{\prime};\B{\theta})\hat{P}(\B{x})}  \right)^{2}.
\end{equation}
The proof that this function satisfies the condition of a symmetric divergence is provided in Appendix \ref{subsec:proof_of_symmetric_divergence}.
When training RBMs, Eq.~\eqref{eq:ratio-divergence} is approximated using the empirical distribution of the data and self-samples, similar to the gradient calculation of forward KLD.

This divergence is designed to incorporate the expectation of both the target distribution $\hat{P}(\B{x})$ and the RBM distribution $P(\B{x}; \theta)$ with equal weights as follows:
\begin{multline}
    \mac{L}(\hat{P}, P;\B{\theta})
    = 2 D_{\mathrm{KL}}[\hat{P} \|P]D_{\mathrm{KL}}[P \|\hat{P}] \\
    + D_{\mathrm{KL}_{2}}[\hat{P} \| P] + D_{\mathrm{KL}_{2}}[P \| \hat{P}],
    \label{eq:ratio-divergence-other-form}
\end{multline}
where
\begin{equation}
    D_{\mathrm{KL}_{2}}[\hat{P}\|P]  \delequal \sum_{\B{x}} \hat{P}(\B{x}) \left(\log \frac{\hat{P}(\B{x})}{P(\B{x};\B{\theta})} \right)^{2},
\end{equation}
which also satisfies the condition of probabilistic divergence.
The first term in Eq.~\eqref{eq:ratio-divergence-other-form}, representing the product of the forward and reverse KLDs, aims to minimize both the forward and reverse KLDs.
The second and third terms are constructed to ensure $\hat{P}(\B{x}) \approx P(\B{x};\B{\theta})$ for samples $\B{x}$ drawn from both $P$ and $\hat{P}$.
This introduces implicit regularization to generate samples efficiently using BGS.
Furthermore, RD, as a divergence defined based on the ratio of probability distributions, does not require computing the target normalizing factors $\hat{Z}$ or that of RBM $Z(\B{\theta})$.

Moreover, this divergence has the following properties.
\begin{theorem}
    For any discrete probability distributions $\hat{P}$ and $P$ parametrized by $\B{\theta}$, the following inequality is satisfied: 
    \begin{equation}
        e^{-(\mac{L}(\hat{P},P;\B{\theta}))^{\nicefrac{1}{2}} }\le \mab{E}_{\hat{P}(\B{x}^{\prime}),P(\B{x};\B{\theta})} \left[ \min \left(1, \frac{\hat{P}(\B{x}^{\prime})P(\B{x}; \B{\theta})}{P(\B{x}^{\prime}; \B{\theta}) \hat{P}(\B{x})} \right) \right].
        \label{eq:inequality_related_with_MH} 
    \end{equation}
\end{theorem}
The detailed derivation is provided in Appendix \ref{subsec:proof_of_inequality_related_with_MH}.
The right-hand side of the inequality can be regarded as the expected value of the acceptance probability in the Metropolis-Hastings algorithm, where $\hat{P}(\B{x})$ represents the target distribution for a state $\B{x}$, and $P(\B{x}; \B{\theta})$ represents the proposal probability for a proposed state $\B{x}^{\prime}$.
Consequently, minimizing $\mac{L}(\hat{P},P;\B{\theta})$ effectively maximizes the acceptance ratio.
In contrast to both the forward and reverse KLD learning, the proposed divergence guarantees a relatively high acceptance rate for the samples generated by a surrogate model.

\section{Related work}\label{sec:related-work}
Extensive research has been conducted on training generative models using target energy functions. 
This section reviews relevant works related to the two main training methods: forward and reverse KLD learning. 
Additionally, we discuss training methods that use other divergences.

\paragraph*{Forward KLD learning}
\citet{mcnaughton2020boosting} and \citet{wu2021unbiased} demonstrated the effectiveness of forward KLD learning in approximating complex probability distributions, even those with multiple modes. However, \citet{ciarella2023machine} pointed out that the forward KLD learning can lead to underfitting the target energy function due to its failure to incorporate it within the learning process. This limitation may impact the efficiency of simulations in various downstream tasks.

\paragraph*{Reverse KLD Learning}
\citet{huang2017accelerated} and \citet{puente2020convolutional} introduced reverse KLD learning in RBMs, significantly enhancing the efficiency of MCMC simulations for various problems. Additionally, \citet{wu2019solving} proposed the reverse KLD learning in Autoregressive Models (ARM), demonstrating its capability to generate samples efficiently from diverse statistical models and the small-scale Sherrington--Kirkpatrick (SK) model without the need for a training dataset. This method also accelerated optimization algorithms for specific discrete optimization problems \citep{hibat2021variational, fan2021finding}. However, \citet{ciarella2023machine} and \citet{inack2022neural} highlighted a significant limitation: in high-dimensional but relatively simple probability distributions with multiple peaks, the reverse KLD learning can lead to mode collapse.

\paragraph*{Other learning methods} 
In addition to forward and reverse KLD learning, several other methods have been proposed. \citet{HYVARINEN20072499} proposed a learning method with a loss function that is a quadratic summation of the probability ratio, similar to our approach. However, unlike our methods, this approach does not incorporate the target energy function, potentially leading to underfitting similar to the forward KLD learning.
Furthermore, \citet{midgley2022flow} introduced $\alpha$-divergence learning, where the divergence approximates the forward KLD as $\alpha \to 1$ and the reverse KLD as $\alpha \to 0$.
In contrast to our methods, $\alpha$-divergence learning requires calculating the normalization factor, which is computationally intensive for MCMC methods in general.

\section{Experiments}\label{sec:experiments}
In this section, we experimentally evaluate the performance of ratio divergence learning across four binary energy-based models: the 2D ferromagnetic Ising model, the Sherrington--Kirkpatrick (SK) model, the maximum independent set (MIS) problem, and the maximum cut problem (MCP).
All experiments included the forward and reverse KLD learning as direct baselines.
Moreover, we add a method in which a loss function is a summation of the forward and reverse KLD to the direct baselines.
Hereafter, this method is referred to as the summation KLD learning.

\subsection{Setting}\label{subsec:setting}

\paragraph*{Evaluation metrics}
To assess the quality of samples generated by the RBMs, we compared the empirical target energy distribution in the training dataset with those generated by the RBM trained using each learning method. 
The Wasserstein metric was used to measure the distance between these empirical distributions quantitatively. 
For the detailed formulation of this metric, see Appendix \ref{subsec:wasserstein}. 
Furthermore, for evaluating the regression performance over the validation dataset $\mac{D}_{\mathrm{val}}$, we define the following error function:
\begin{subequations}
    \label{eq:mse_of_energy_difference}
    \begin{gather}
        R(\theta) = \frac{1}{|\mac{D}_{\mathrm{val}}|^2} \sum_{\B{x}',\B{x}\in\mac{D}_{\mathrm{val}}} \left( \Delta F(\B{x}',\B{x};\B{\theta}) - \Delta \hat{E}(\B{x}',\B{x}) \right)^2,\\
        \Delta F(\B{x}',\B{x};\B{\theta}) = F(\B{x}';\B{\theta}) - F(\B{x};\B{\theta}),\\
        \Delta \hat{E}(\B{x}',\B{x}) = \hat{E}(\B{x}') - \hat{E}(\B{x}).
    \end{gather}
\end{subequations}
At first glance, it seems that the average of $F(\B{x};\B{\theta}) - \hat{E}(\B{x})$ is sufficient for evaluating regression performance.
However, in most downstream tasks, the energy difference between two states is more important than the absolute value of the energy itself.
Therefore, we introduce Eq.~\eqref{eq:mse_of_energy_difference} as a mean squared error of the energy difference between two states.

We also conducted a principal component analysis (PCA) on both the generated samples and the training dataset to investigate mode collapse.
We then visualized the generated samples in the two-dimensional space defined by the dominant principal components. Furthermore, we analyze the empirical distribution defined by $\{\nicefrac{d_{H}(\B{x}{\mu}, \B{x}{\nu})}{N_x}\}_{\mu<\nu}^K$, where $d_{H}$ denotes the Hamming distance and $K$ is the number of samples. This approach is commonly used as a diversity measure in combinatorial optimization problems \citep{hanaka2023framework, fernau2019algorithmic}.

\paragraph*{Datasets}
In all experiments, the training dataset, comprising 16,384 samples, and the validation dataset, comprising 1,024 samples, were generated using the exchange MCMC simulations \citep{Hukushima_1996_J.Phys.Soc.Jpn.}.
The exchange MCMC simulations were conducted for a sufficient duration to ensure sampling from the stationary distributions of these target energy-based models. 
For further details on these simulations, refer to Appendix \ref{subsec:pt_setting}.

\paragraph*{Implementation}
The number of hidden variables is set to be equal to the number of visible variables for each target energy-based model. 
The size of the self-sample set is set equal to that of the training dataset.
The Adam optimizer is employed with the following configurations: $\mathrm{lr} = 0.001$, $\beta_1 = 0.9$, $\beta_2 = 0.999$, and $\epsilon = 1.0 \times 10^{-8}$, with the minibatch size set to 128~\citep{Kingma_2014_arXiv.org}. 
The model is trained for up to $1{,}000$ epochs. 
After training the RBMs, we generated $16{,}384$ samples using BGS with 100 steps, setting the training dataset as the initial condition. 
Training and sampling were performed five times with different random seeds. To evaluate the empirical distribution of the Hamming distance, we randomly extracted 1,000 samples from the samples generated by the RBMs and from the training dataset.
All experiments were conducted on an Intel(R) Xeon(R) Gold 6240 CPU @ 2.60GHz with a Tesla V100.

\subsection{2D Ising model}\label{subsec:2d-ising}
\begin{figure*}[t]
    \centering
    \begin{minipage}{0.495\linewidth}
        \centering
        \begin{subcaptiongroup}
            \begin{minipage}[t]{0.49\linewidth}
                \centering
                \phantomcaption
                \label{fig:ising_riw}
                \begin{overpic}[width=\textwidth]{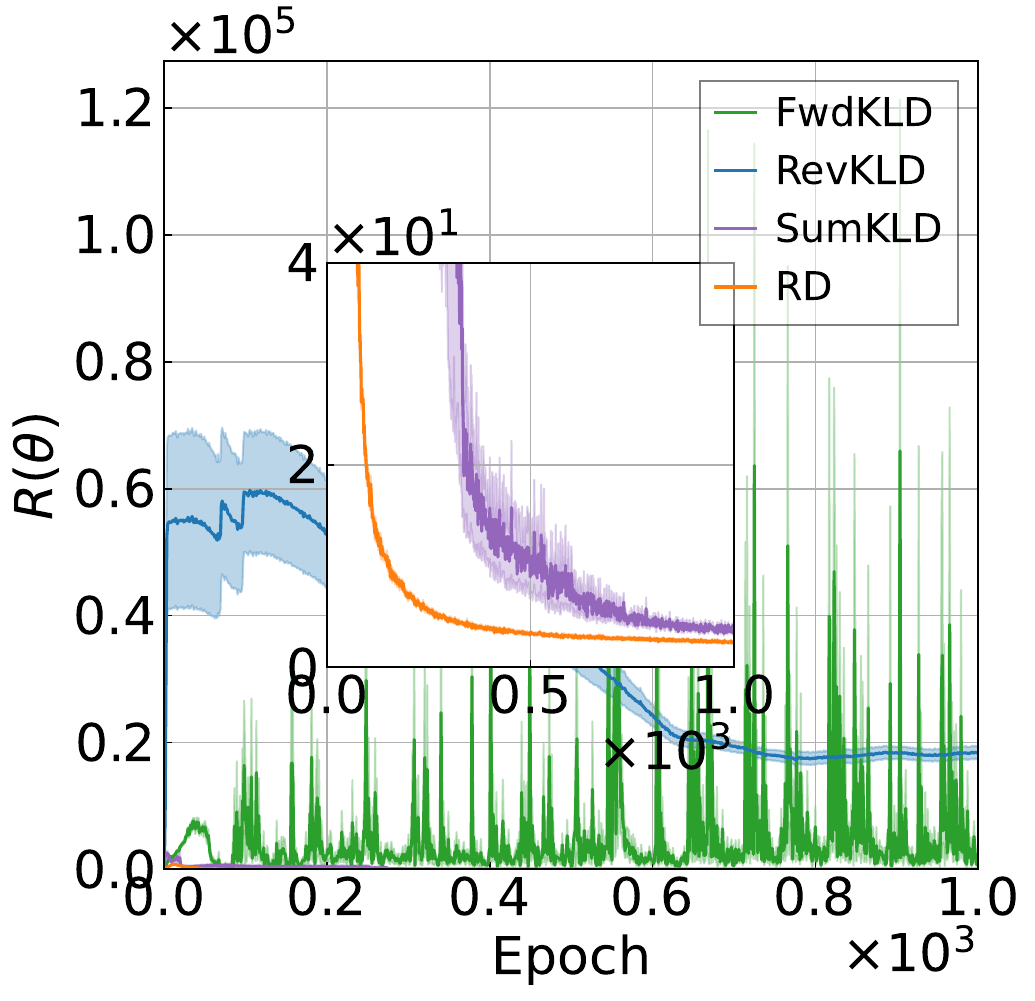}
                    \put(0,95){\subfiglabelsize\captiontext*{}}
                \end{overpic}
            \end{minipage}
            \hfill
            \begin{minipage}[t]{0.49\linewidth}
                \centering
                \phantomcaption
                \label{fig:ising_energy_dist}
                \begin{overpic}[width=\textwidth]{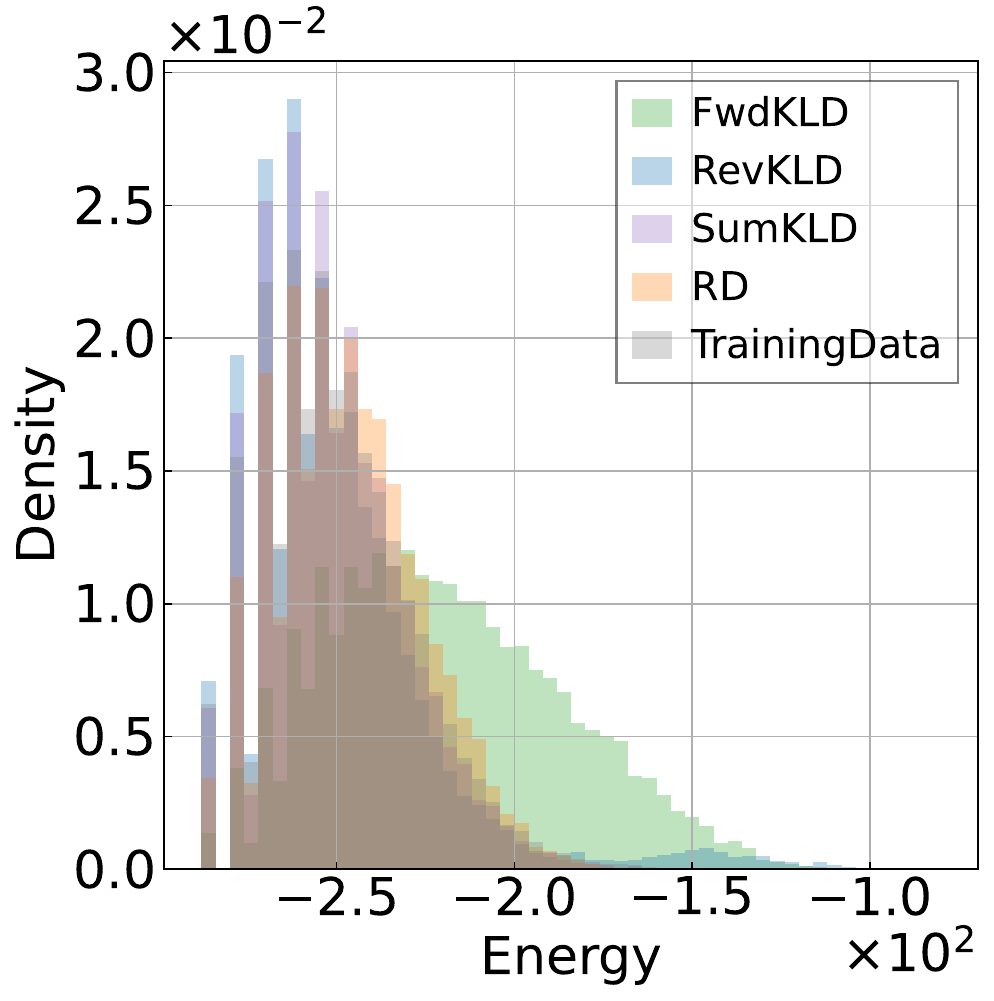}
                    \put(0,95){\subfiglabelsize\captiontext*{}}
                \end{overpic}
            \end{minipage}
        \end{subcaptiongroup}
        \caption{\subref{fig:ising_riw} $R(\theta)$ as a function of epochs during the training process, and \subref{fig:ising_energy_dist} empirical energy distributions of the generated samples by each method compared to those in the training dataset for the Ising model on a $12 \times 12$ square lattice.}
        \label{fig:ising_riw_and_energy_dist}
    \end{minipage}
    \begin{minipage}{0.495\linewidth}
        \centering
        \begin{subcaptiongroup}
            \begin{minipage}[t]{0.49\linewidth}
                \centering
                \phantomcaption
                \label{fig:ising_dependeny_of_riw_on_size}
                \begin{overpic}[width=\textwidth]{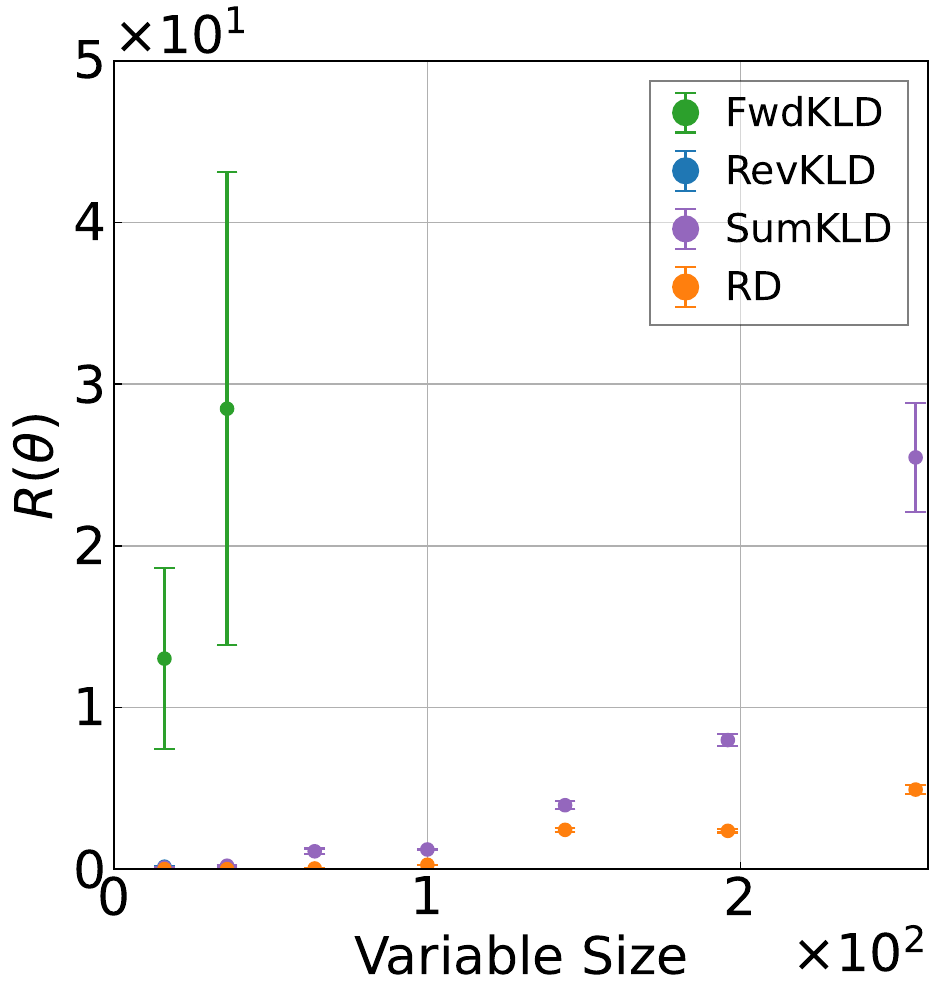}
                    \put(0,95){\subfiglabelsize\captiontext*{}}
                \end{overpic}
            \end{minipage}
            \hfill
            \begin{minipage}[t]{0.49\linewidth}
                \centering
                \phantomcaption
                \label{fig:ising_dependeny_of_wasserstein_on_size}
                \begin{overpic}[width=\textwidth]{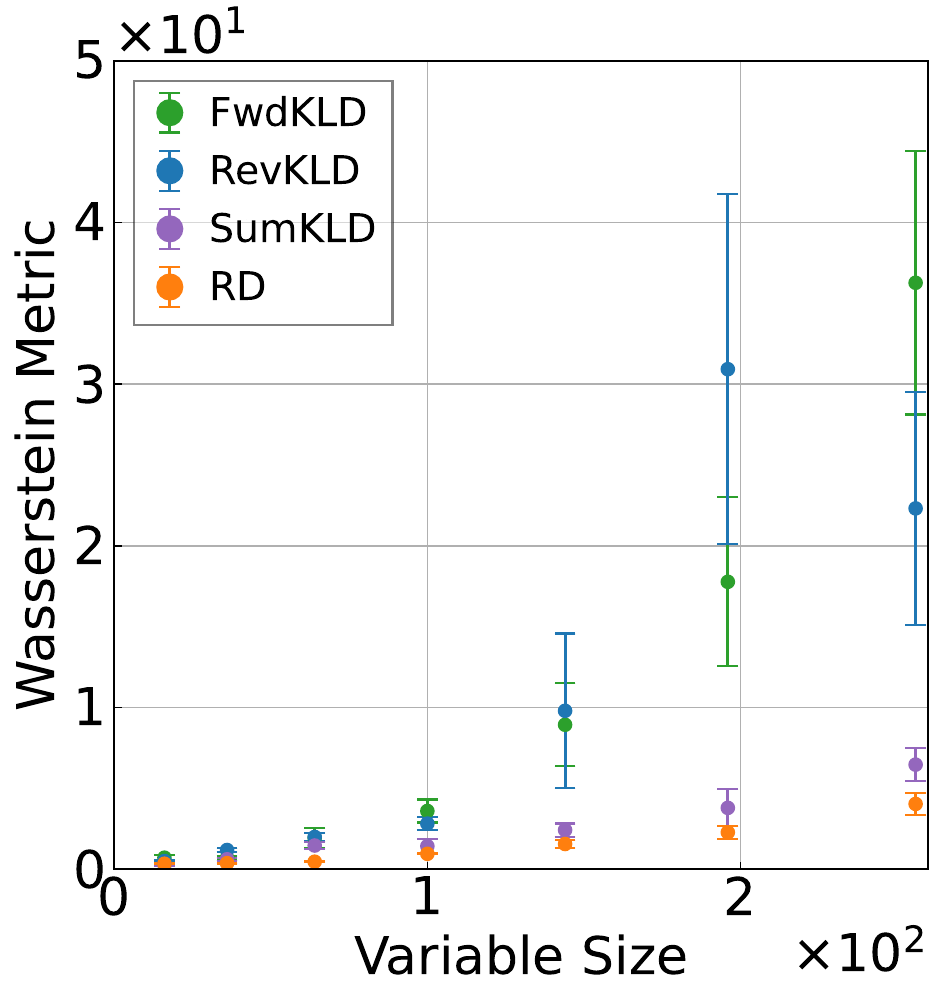}
                    \put(0,95){\subfiglabelsize\captiontext*{}}
                \end{overpic}
            \end{minipage}
        \end{subcaptiongroup}
        \caption{\subref{fig:ising_dependeny_of_riw_on_size} $R(\theta)$ and \subref{fig:ising_dependeny_of_wasserstein_on_size} Wasserstein metric of each method as a function of the dimension of the 2D Ising models. The error bars represent the standard error.}
        \label{fig:ising_dependeny_on_size}
    \end{minipage}
\end{figure*}

To demonstrate the performance gaps between RD learning and KLD learning, even for a simple distribution, we consider the Ising model on a 2D square lattice graph, denoted as $G(V, E)$, whose energy function is defined by $\hat{E}(\B{x}) = - J \sum_{(i, j) \in E} x_{i} x_{j}$, where $\B{x} \in \{-1, +1\}^{N_x}$ denotes the state configuration and $J$ is a positive scalar representing the interaction strength between neighboring variables. 
This is the minimal model for ferromagnetic materials that exhibit a \textit{phase transition}. 
As $\beta$ increases, the data distribution transitions from non-clustered to two-clustered states. 
Here, we set the graph $G$ to a $12 \times 12$ square lattice graph with periodic boundary conditions and $J$ to $1$.
We show the performance of RBM at $\beta = 0.5$, corresponding to the two clustered states. 
The results are summarized as follows.

Figure~\ref{fig:ising_riw} shows the evolution of $R(\theta)$ as functions of epochs during the training process.
RD learning not only converges faster than the other learning methods by $1000$ epochs but also demonstrates significantly more stable learning process over the epochs.
This stability is one of the key advantages of RD learning. 
Moreover, Figure~\ref{fig:ising_energy_dist} compares the empirical distribution of the generated samples with the target energy distributions from the training dataset, showing the accuracy of the learned distributions after training.
The Wasserstein metric for RD learning is $1.6$, which is considerably lower than that of the forward KLD learning at $8.9$, the reverse KLD learning at $9.8$, and the summation KLD learning at $2.4$.
These results indicate that RD learning overcomes the other methods in terms of sampling performance.

We also examine the dimension-dependence of $R(\theta)$ at $1000$ epochs in Figure~\ref{fig:ising_dependeny_of_riw_on_size}, and the Wasserstein metric in Figure~\ref{fig:ising_dependeny_of_wasserstein_on_size}, for each learning method. 
Across different number of variables, RD learning outperforms the other methods. 
Notably, when employing RD learning, the rate of increase in the number of variables for $R(\theta)$ and the Wasserstein metric is lower, underscoring its efficiency compared to the other methods.
These results indicate that RD learning achieves both high energy regression accuracy and effective mode covering of the distribution.

\begin{table}[t]
\caption{Wasserstein metric between the empirical target energy distributions of the samples generated by RBMs, trained using each learning method, and the distribution of the training dataset. The model sizes are indicated in parentheses.}
\label{tab:wasserstein_metric}
\begin{center}
\footnotesize
\scshape
\begin{tabular}{lcccc}
\hline
 & FwdKLD & RevKLD & SumKLD & RD \\
\hline
Ising ($144$) & $8.9 \pm 2.6$ & $9.8 \pm 4.9$ & $2.4 \pm 0.4$ & $\mathbf{1.6 \pm 0.2}$ \\
SK ($144$) & $1.9 \pm 0.5$ & $1.5 \pm 1.5$ & $1.1 \pm 0.7$ & $\mathbf{0.5 \pm 0.1}$  \\
MIS ($250$) & $\mathbf{2.0 \pm 0.2}$ & $57.0 \pm 18.7$ & $5.7 \pm 0.3$ & $4.8 \pm 0.7$  \\
Gset G1 ($800$) & $34.6 \pm 1.8$ & $85.9 \pm 7.9$ & $\mathbf{22.7 \pm 9.0}$ & $619.1 \pm 36.1$  \\
Gset G6 ($800$) & $13.3 \pm 2.1$ & $182.9 \pm 9.8$ & $\mathbf{5.7 \pm 0.7}$& $124.6 \pm 39.6$  \\
Gset G14 ($800$) & $\mathbf{20.5 \pm 1.5}$ & $67.0 \pm 3.2$ & $31.7 \pm 8.5$ & $66.0 \pm 33.4$ \\
Gset G18 ($800$) & $17.8 \pm 2.2$ & $92.5 \pm 5.5$ & $\mathbf{5.9 \pm 0.4}$ & $9.1 \pm 0.7$ \\
\hline
\end{tabular}
\end{center}
\end{table}

\subsection{Sherrington--Kirkpatrick model}\label{subsec:sk-model}
Next, to demonstrate the advantages of RD learning with a more complex model than the 2D Ising model, we evaluate the performance on the SK model, defined by the energy function $\hat{E}(\B{x}) = - \sum_{i<j} J_{ij} x_i x_j$, where $\B{x} \in \{-1, +1\}^{N_x}$ represents the state configuration and $J_{ij}$ represents interactions drawn from an independent and identically distributed normal distribution $\mathcal{N}(0, \nicefrac{1}{N_x})$.
Unlike the Ising model, this model is defined on a complete graph, where every pair of distinct nodes is connected.
This randomness introduces a large number of peaks in the state space $\mathcal{X}$ for large $\beta$. 
We set $N_x=144$ and $\beta=2.0$, where the SK model exhibits multi-clustered states.
These experiments used five sets of $\{J_{ij}\}$ values.

\begin{figure*}[t]
    \centering
    \begin{minipage}{0.495\linewidth}
        \centering
        \begin{subcaptiongroup}
            \begin{minipage}[t]{0.49\linewidth}
                \centering
                \phantomcaption
                \label{fig:sk_model_riw}
                \begin{overpic}[width=\textwidth]{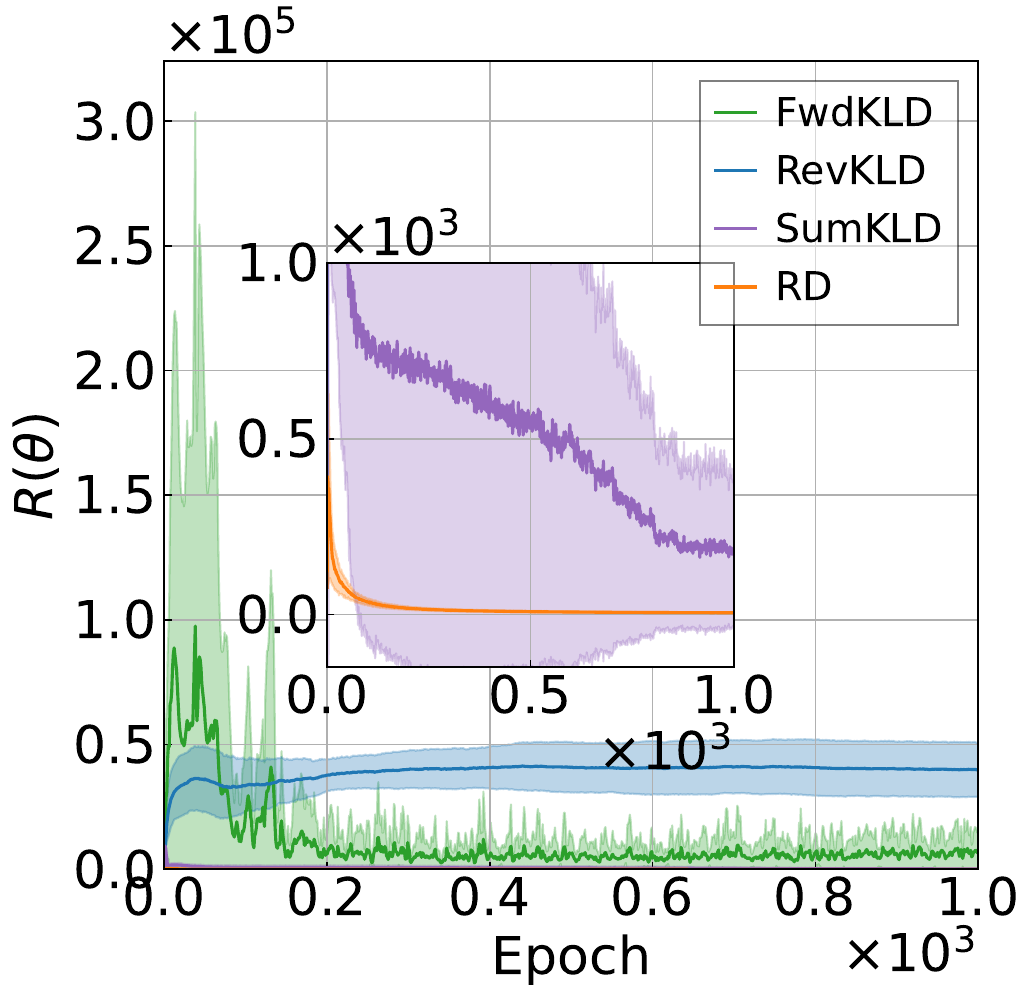}
                    \put(0,95){\subfiglabelsize\captiontext*{}}
                \end{overpic}
            \end{minipage}
            \hfill
            \begin{minipage}[t]{0.49\linewidth}
                \centering
                \phantomcaption
                \label{fig:sk_model_energy_dist}
                \begin{overpic}[width=\textwidth]{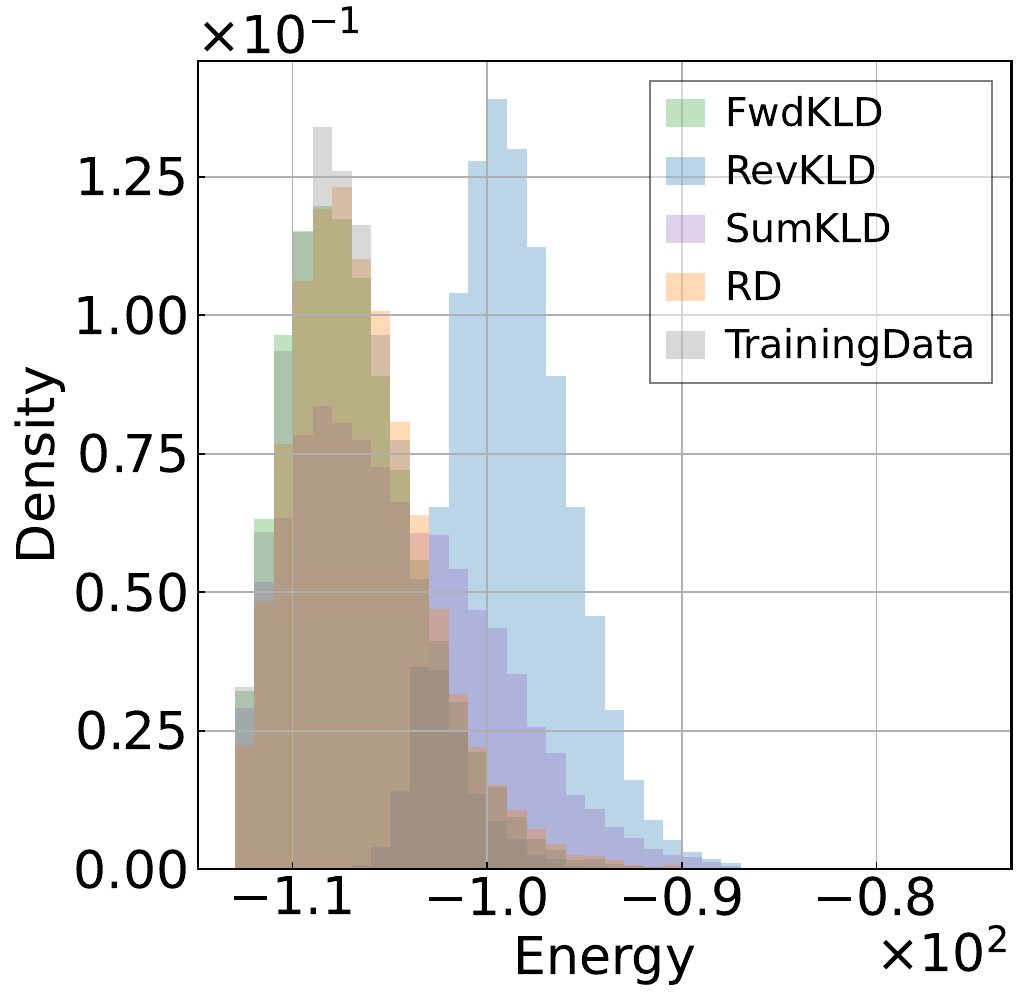}
                    \put(0,95){\subfiglabelsize\captiontext*{}}
                \end{overpic}
            \end{minipage}
        \end{subcaptiongroup}
        \caption{\subref{fig:sk_model_riw} $R(\theta)$ as functions of epochs during the training process, and \subref{fig:sk_model_energy_dist} empirical energy distributions of the generated samples by each method and of those in the training dataset on the SK model with $N_x=144$.}
        \label{fig:sk_model_riw_and_energy_dist}
    \end{minipage}
    \begin{minipage}{0.495\linewidth}
        \centering
        \begin{subcaptiongroup}
            \begin{minipage}[t]{0.49\linewidth}
                \centering
                \phantomcaption
                \label{fig:sk_model_depenency_of_riw_on_size}
                \begin{overpic}[width=\textwidth]{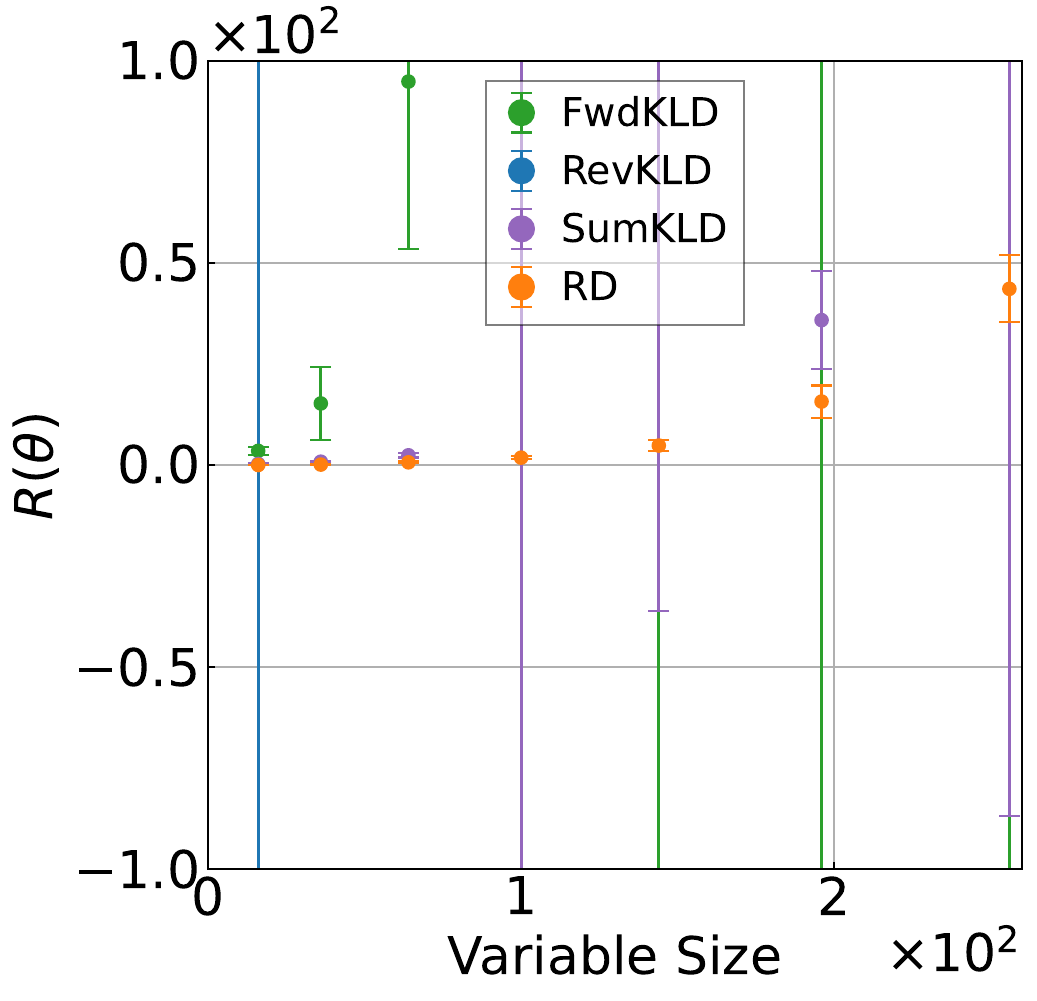}
                    \put(0,95){\subfiglabelsize\captiontext*{}}
                \end{overpic}
            \end{minipage}
            \hfill
            \begin{minipage}[t]{0.49\linewidth}
                \centering
                \phantomcaption
                \label{fig:sk_model_depenency_of_wasserstein_on_size}
                \begin{overpic}[width=\textwidth]{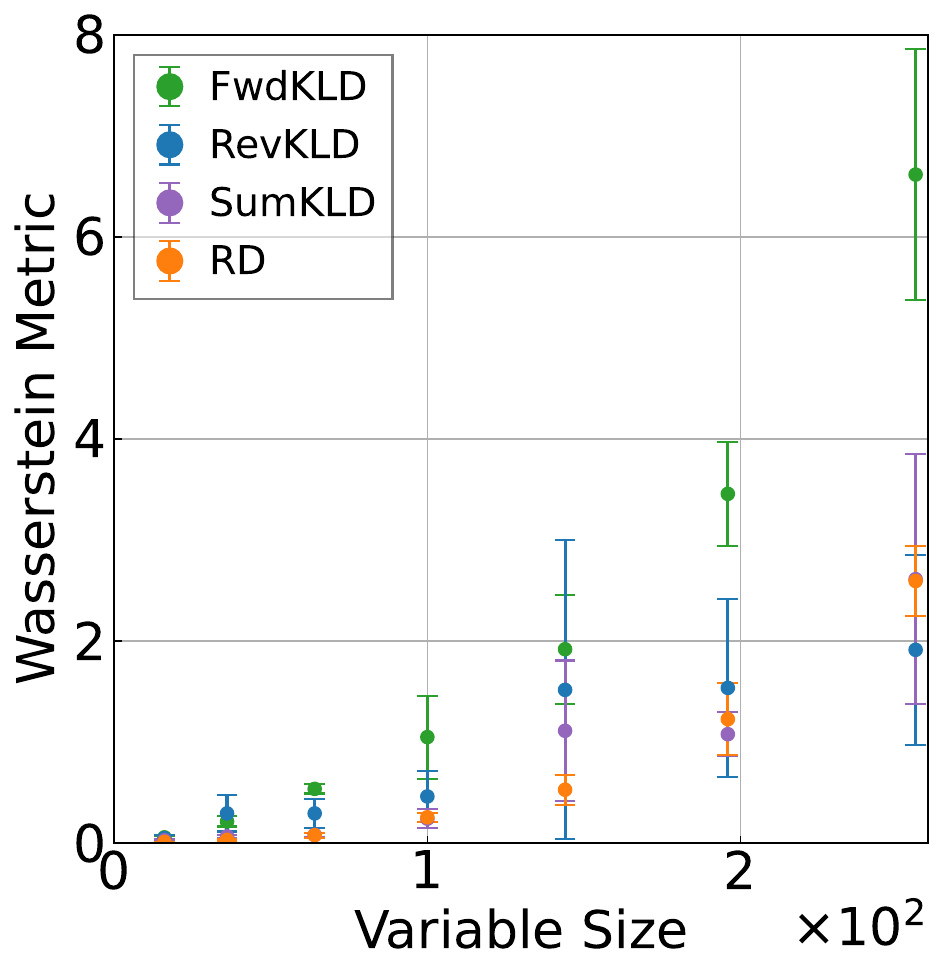}
                    \put(0,95){\subfiglabelsize\captiontext*{}}
                \end{overpic}
            \end{minipage}
        \end{subcaptiongroup}
        \caption{\subref{fig:sk_model_depenency_of_riw_on_size} $R(\theta)$ and \subref{fig:sk_model_depenency_of_wasserstein_on_size} Wasserstein distance of each method as a function of the dimension of the SK models. The error bar represents the standard deviation of different graphs.}
        \label{fig:sk_model_depenency_on_size}
    \end{minipage}
\end{figure*}

Figure~\ref{fig:sk_model_riw} demonstrates that RD learning outperforms the other methods at $1000$ epochs in terms of $R(\theta)$.
Moreover, RD learning exhibits significantly smaller error bars for $R(\theta)$, demonstrating robust performance across a variety of problems. In contrast, summation divergence learning shows larger errors in $R(\theta)$, which can lead to substantial performance degradation depending on the problem. 
As shown in Figure~\ref{fig:sk_model_energy_dist}, the empirical energy distribution of the samples generated by the RBM trained using RD learning more accurately replicates the target energy distribution in the training dataset.
See Table~\ref{tab:wasserstein_metric} for the value of the Wasserstein metric.
Furthermore, Figure~\ref{fig:sk_model_depenency_on_size} shows that RD learning consistently outperforms other learning methods in terms of $R(\theta)$ across different numbers of variables. 
The rate of increase in the $R(\theta)$ metric for RD learning is significantly lower than other methods.
Furthermore, the increase in the Wasserstein metric for RD learning is comparable to that of summation and reverse KLD learning. 
These results indicate that RD learning achieves both high energy regression and effective mode covering of the distribution.

\begin{figure}[t]
    \centering
    \centering
    \begin{subcaptiongroup}
        \begin{minipage}[t]{0.49\linewidth}
            \centering
            \phantomcaption
            \label{fig:mis_riw}
            \begin{overpic}[width=\linewidth]{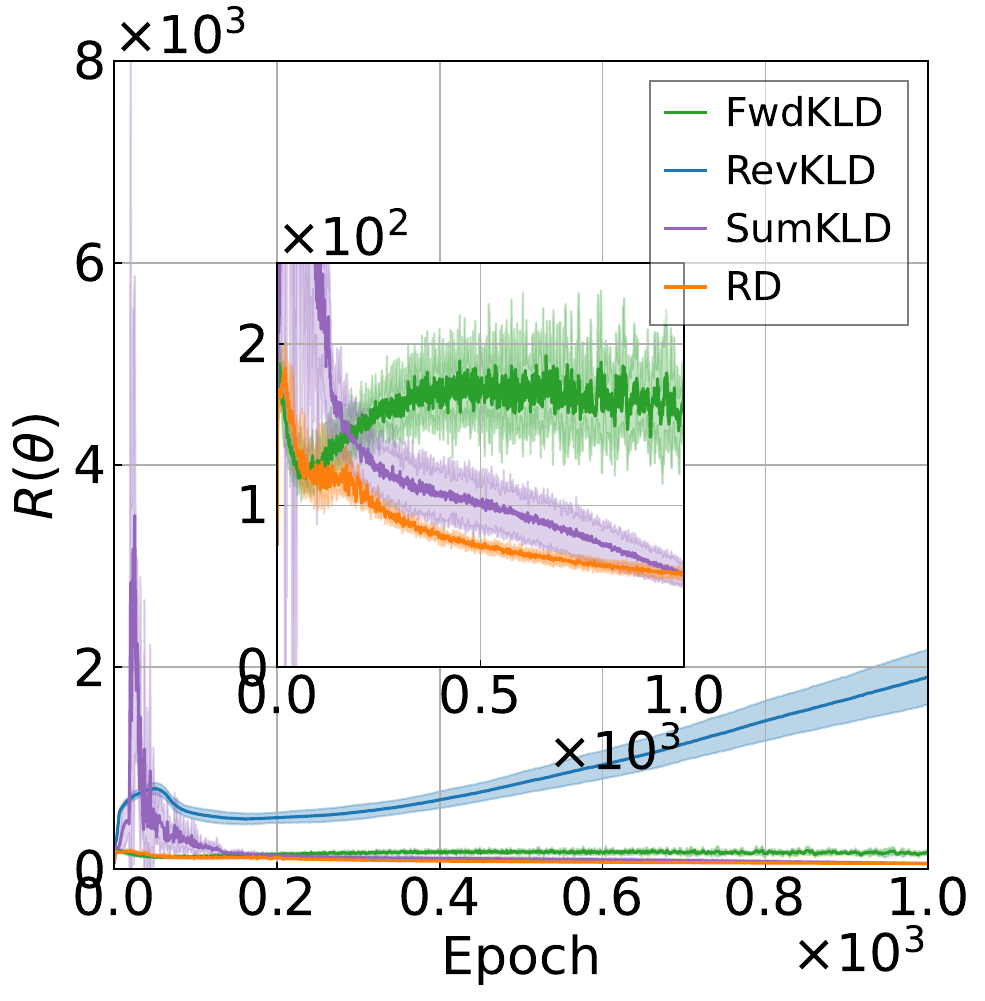}
                \put(0,95){\subfiglabelsize\captiontext*{}}
            \end{overpic}
        \end{minipage}
        \hfill
        \begin{minipage}[t]{0.49\linewidth}
            \centering
            \phantomcaption
            \label{fig:mis_energy_dist}
            \begin{overpic}[width=\textwidth]{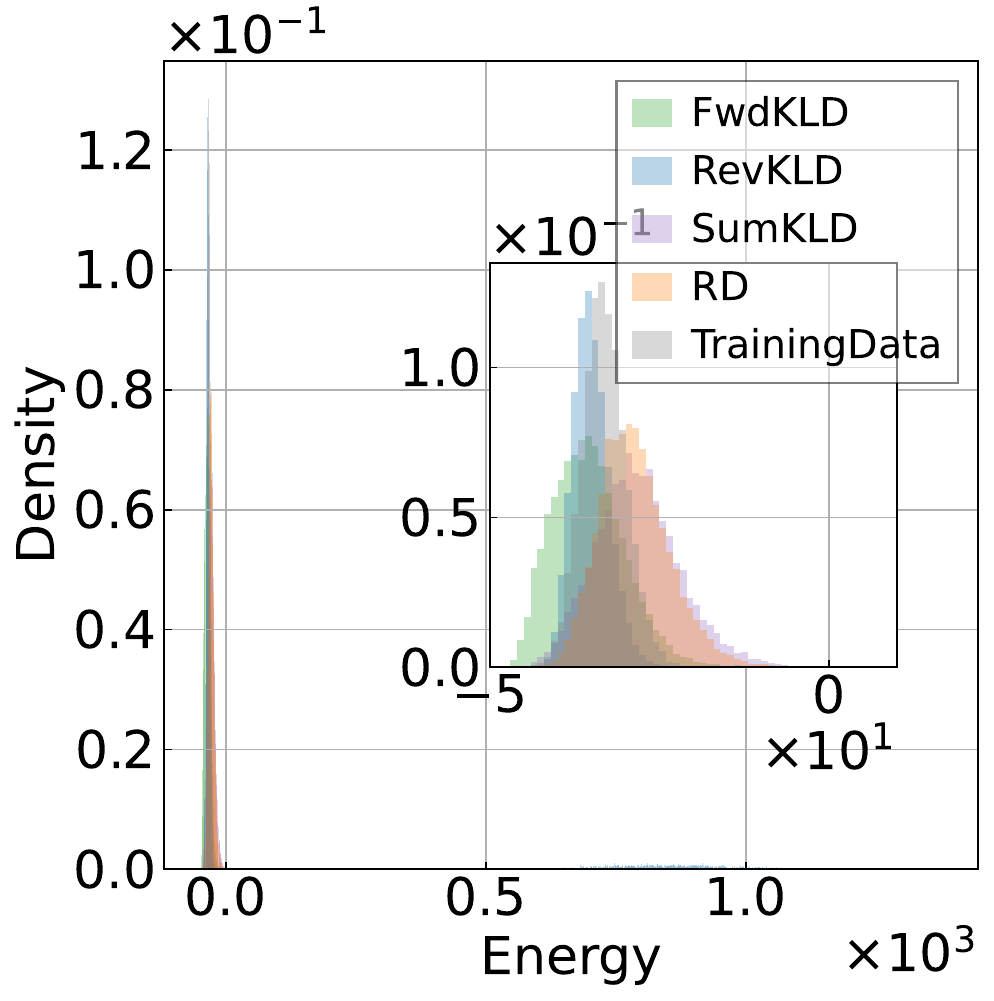}
                \put(0,95){\subfiglabelsize\captiontext*{}}
            \end{overpic}
        \end{minipage}
    \end{subcaptiongroup}
    \caption{\subref{fig:mis_riw} $R(\theta)$ as a function of epochs during the training process of RBMs, and \subref{fig:mis_energy_dist} empirical energy distributions of samples generated by each model compared to those in the training dataset on the MIS of a regular random graph with 250 nodes and a degree of 20.}
    \label{fig:mis_riw_and_energy_dist}
\end{figure}

\subsection{Maximum independent set}
Next, we consider the MIS problem, which involves finding the largest possible independent set in a given graph $G(V, E)$ and is noted by \citet{ciarella2023machine} that approximating the distribution is difficult.
The energy function for this problem consists of two terms: the size of the independent set and a penalty term, defined as $\hat{E}(x) = - \sum_{i \in V} x_i + \alpha \sum_{(i,j) \in E} x_i x_j$, where $x_i = {0,1}$ indicates whether node $i$ is part of the independent set, and $\alpha$ represents the penalty strength.
The MIS problem is characterized by an exponential number of local optima. Consequently, this study uses the MIS problem to evaluate whether RD learning can capture the exponential number of peaks in this distribution. The experiments utilize regular random graphs with $250$ nodes and a degree of $20$, analyzing the performance of the RBM at $\alpha = 2$ and $\beta = 2.0$, corresponding to a distribution with an exponential number of peaks.
The experiments were conducted on five different graphs.

\begin{figure}[t]
    \centering
    \begin{subcaptiongroup}
        \begin{minipage}[t]{0.49\linewidth}
            \centering
            \phantomcaption
            \label{fig:mis_pca_fwdkl}
            \begin{overpic}[width=\linewidth]{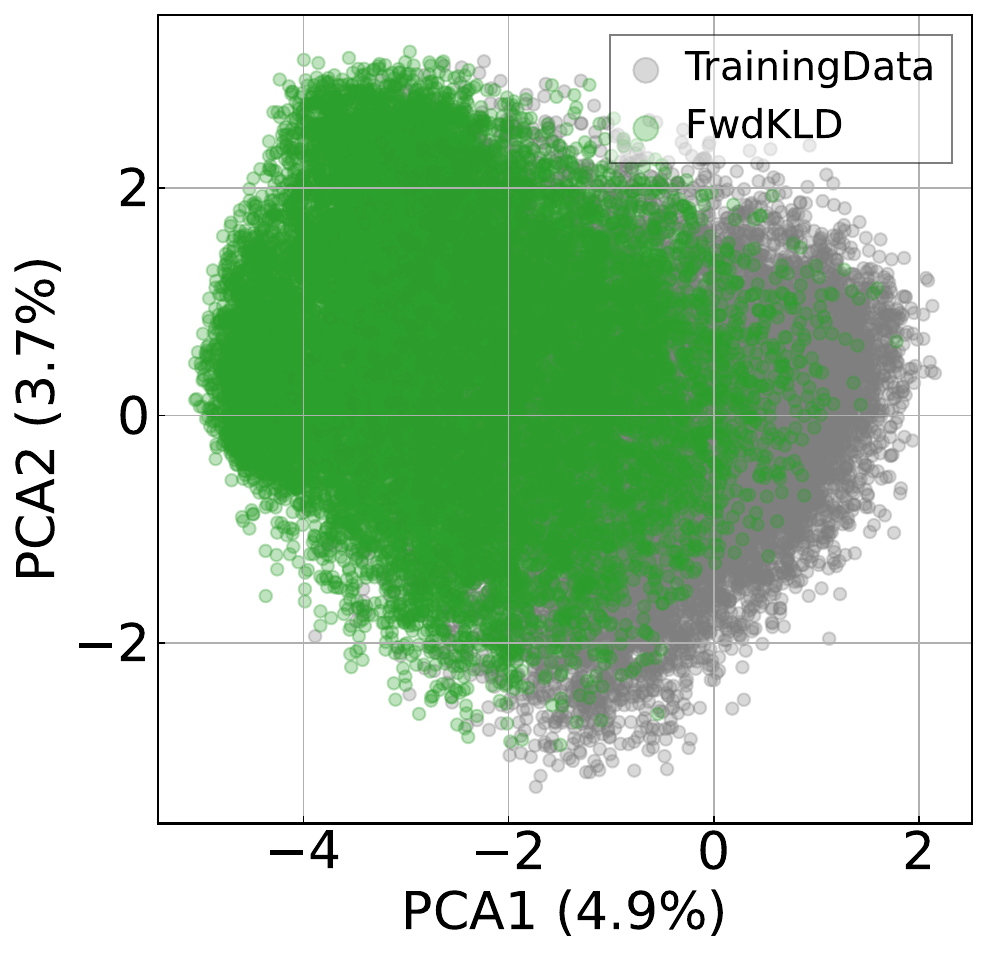}
                \put(0,95){\subfiglabelsize\captiontext*{}}
            \end{overpic}
        \end{minipage}
        \hfill
        \begin{minipage}[t]{0.49\linewidth}
            \centering
            \phantomcaption
            \label{fig:mis_pca_revkl}
            \begin{overpic}[width=\textwidth]{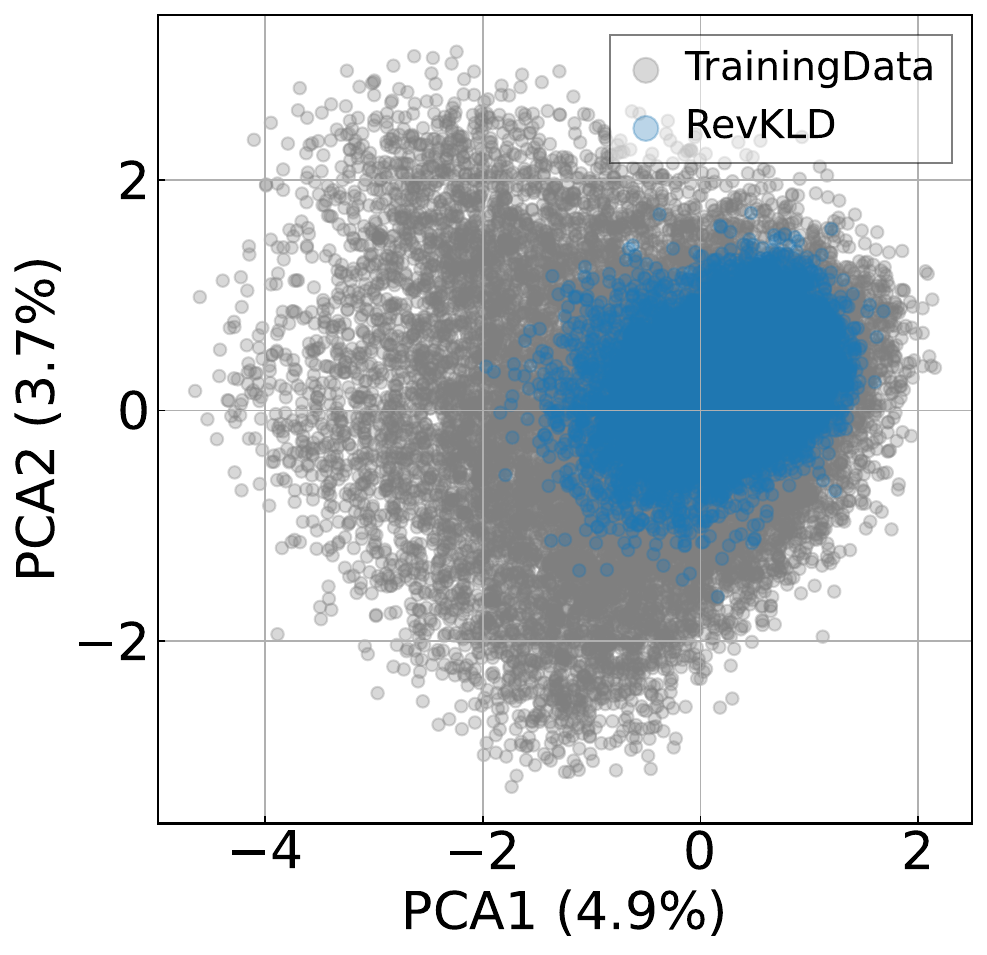}
                \put(0,95){\subfiglabelsize\captiontext*{}}
            \end{overpic}
        \end{minipage}
        \hfill
        \begin{minipage}[t]{0.49\linewidth}
            \centering
            \phantomcaption
            \label{fig:mis_pcs2d_fwdrevkl}
            \begin{overpic}[width=\textwidth]{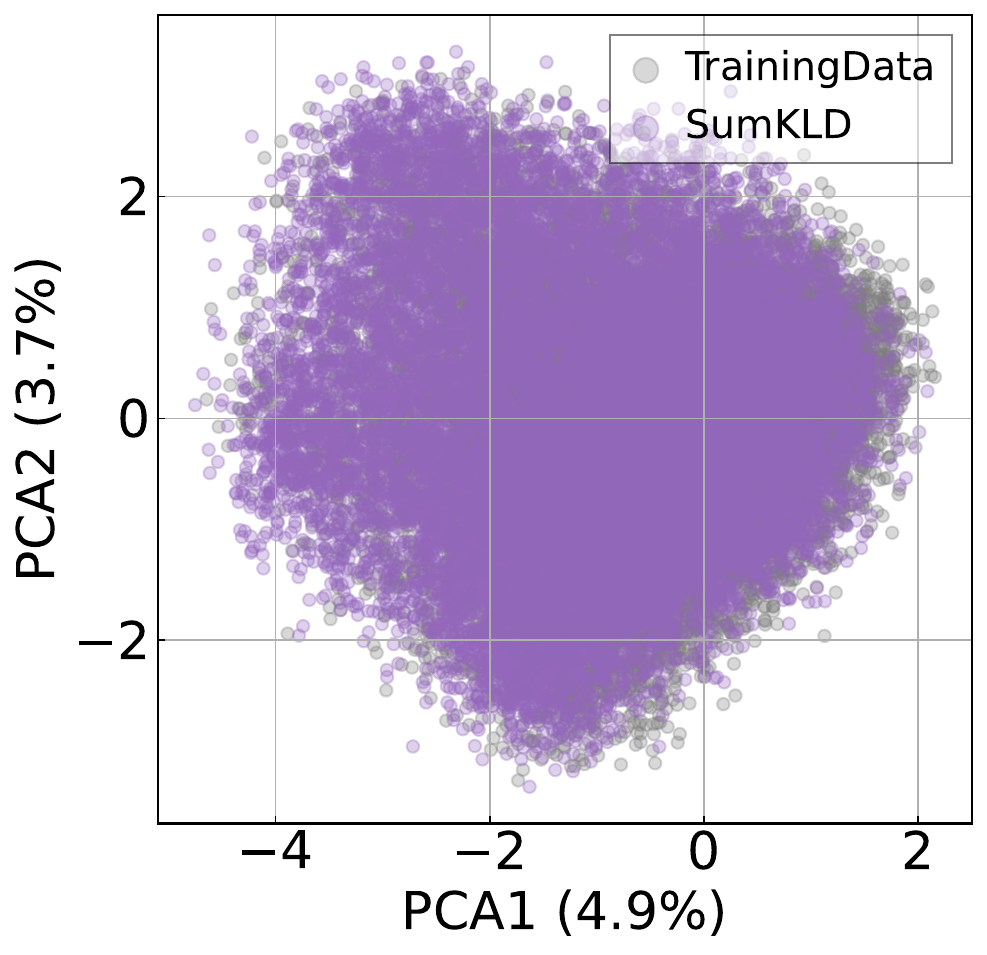}
                \put(0,95){\subfiglabelsize\captiontext*{}}
            \end{overpic}
        \end{minipage}
        \hfill
        \begin{minipage}[t]{0.49\linewidth}
            \centering
            \phantomcaption
            \label{fig:mis_pca_rd}
            \begin{overpic}[width=\linewidth]{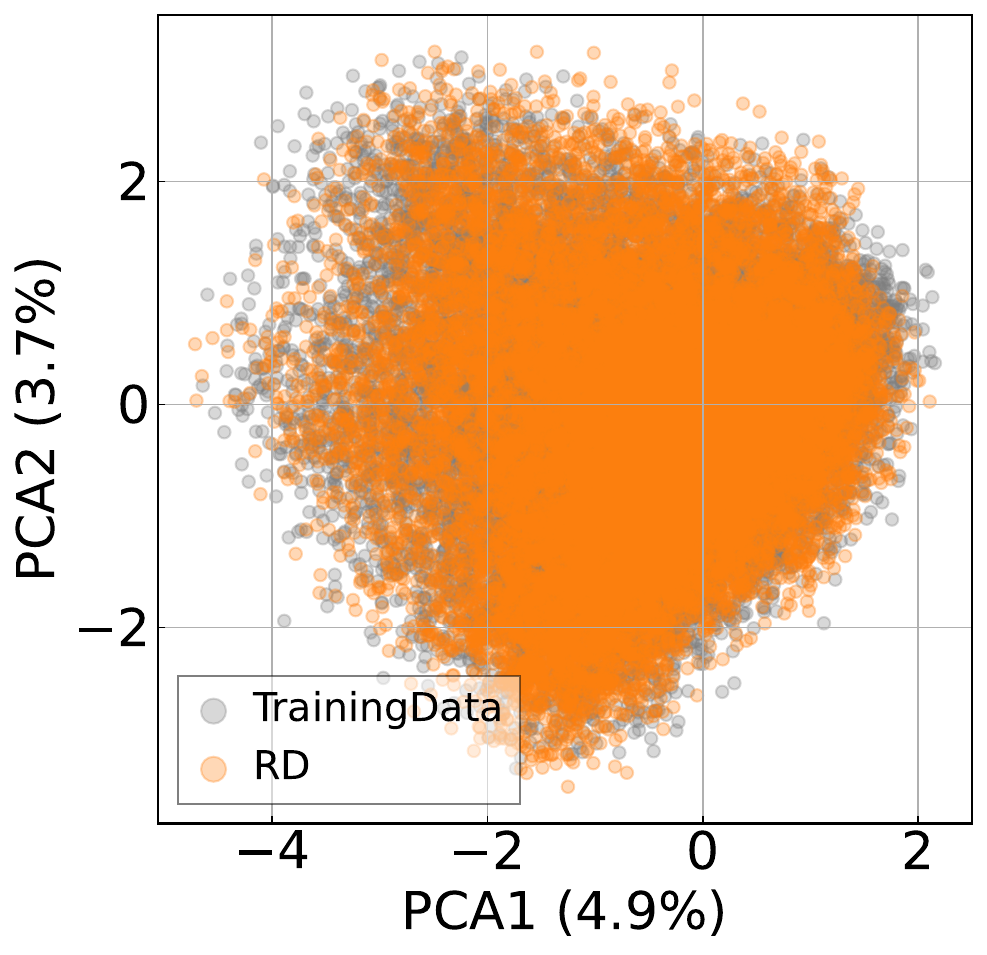}
                \put(0,95){\subfiglabelsize\captiontext*{}}
            \end{overpic}
        \end{minipage}
    \end{subcaptiongroup}
    \caption{\subref{fig:mis_pca_fwdkl}--\subref{fig:mis_pca_rd} Two dimensional PCA mapping of generated samples by each method for the MIS on a regular random graph with 250 nodes and a degree of 20.}
    \label{fig:mis_pca}
\end{figure}

\begin{figure}[t]
    \centering
    \includegraphics[width=\linewidth]{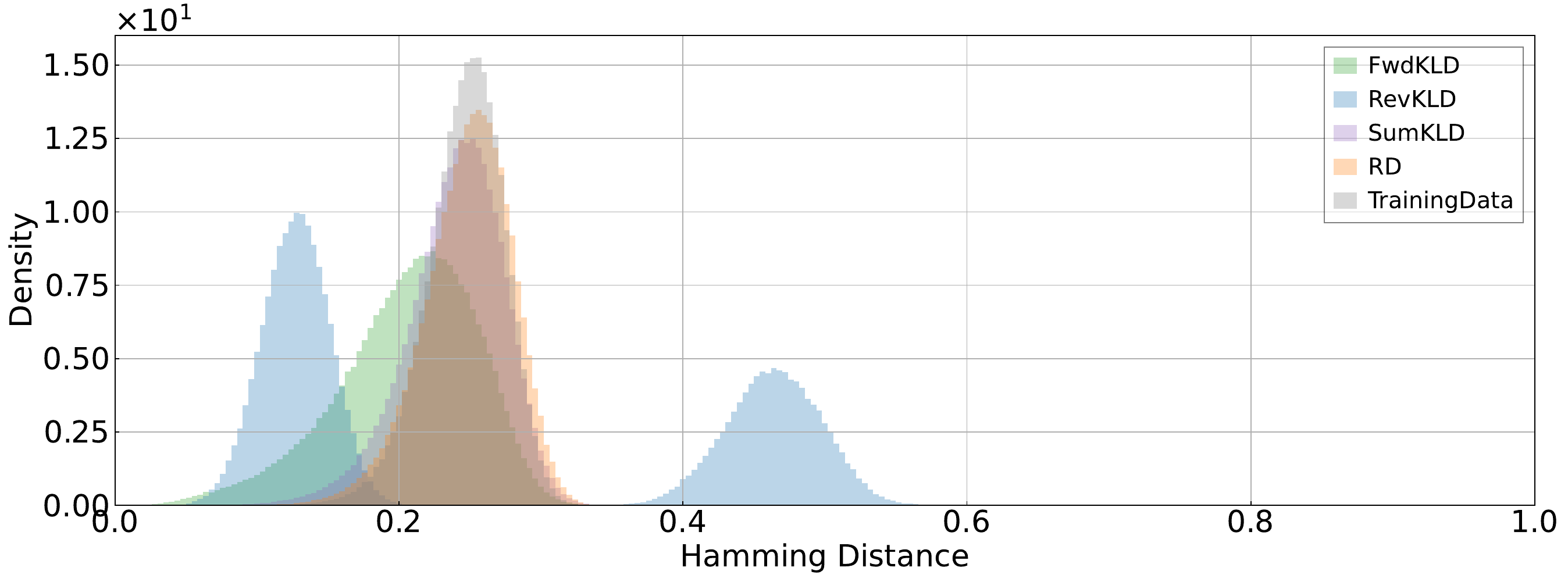}
    \caption{Empirical distribution of the hamming distance of the generated samples by each method and those in the training dataset.}
    \label{fig:mis_hamming_dist}
\end{figure}

Figure~\ref{fig:mis_riw} demonstrates that RD learning significantly outperforms both the forward and reverse KLD learning in terms of $R(\theta)$.
Additionally, RD learning shows a consistent decrease in $R(\theta)$.
In contrast, the $R(\theta)$ curves for forward and reverse KLD learning initially decrease but then increase, suggesting that these methods lead to overfitting of the target distribution.
This result implies the presence of implicit regularization in RD learning.
Although the summation KLD learning has these features, the performance of RD learning is more robust to changes in graph structures compared to the summation KLD learning, as evidenced by the smaller shaded area representing the standard deviation.
As shown in Figure~\ref{fig:mis_energy_dist}, the sampling performance of RD learning is comparable to that of the forward KLD learning, with the Wasserstein metric for RD learning at $4.8$ being less than $2.5$ times that of the $2.0$ achieved by the forward KLD learning.
Additionally, Figures~\ref{fig:mis_pca_fwdkl}--\ref{fig:mis_pca_rd} present two-dimensional maps of generated samples using PCA.
These figures illustrate that the forward and reverse KLD learning cause the RBMs to generate samples only within a limited portion of the region occupied by the training dataset.
In contrast, RD learning and the summation KLD learning enable the RBM to generate samples uniformly across the entire region.
Furthermore, Figure~\ref{fig:mis_hamming_dist} demonstrates that the Hamming distance of samples generated by the model trained with RD learning closely matches that of the training dataset samples, outperforming the other learning methods.
These results indicate that RD learning achieves both high energy regression and effective mode covering of the distribution.


\subsection{Maximum cut problems}
An MCP is a discrete optimization problem that aims to find the cut that maximizes the total weight of edges between two partitions of a given graph. The energy function is defined as $\hat{E}(\B{x}) = - \sum_{i<j} w_{ij} (x_i - x_j)^2$, where $w_{ij}$ is the weight of edge $(i,j)$, and $x_i = {0, 1}$ indicates the side of the cut on which node $i$ is located.
We evaluate our method on various heterogeneous graph topologies using MCPs: G1, G6, G14, and G18 from the Gset benchmark~\cite{Gset} as practical examples.
G1 is a random graph with 800 nodes and a density of 6\%, where all weights are set to $+1$.
G6 has the same structure as G1 but with edge weights of either $+1$ or $-1$.
G14 is a union graph composed of two random planar graphs with 800 nodes and a density of 99\%, where all weights are set to $+1$.
G18 has the same structure as G14 but with weights of either $+1$ or $-1$.
The results for G1 are presented in the main text, while additional results can be found in Appendix~\ref{sec:gset_additional_result}.
In this study, we analyze the performance of the RBM at $\beta = 1.0$, which corresponds to the conditions under which an optimal solution is typically found using the exchange Monte Carlo method within a realistic time frame.

\begin{figure}[t]
    \centering
    \begin{subcaptiongroup}
        \begin{minipage}[t]{0.49\linewidth}
            \centering
            \phantomcaption
            \label{fig:g1_riw}
            \begin{overpic}[width=\textwidth]{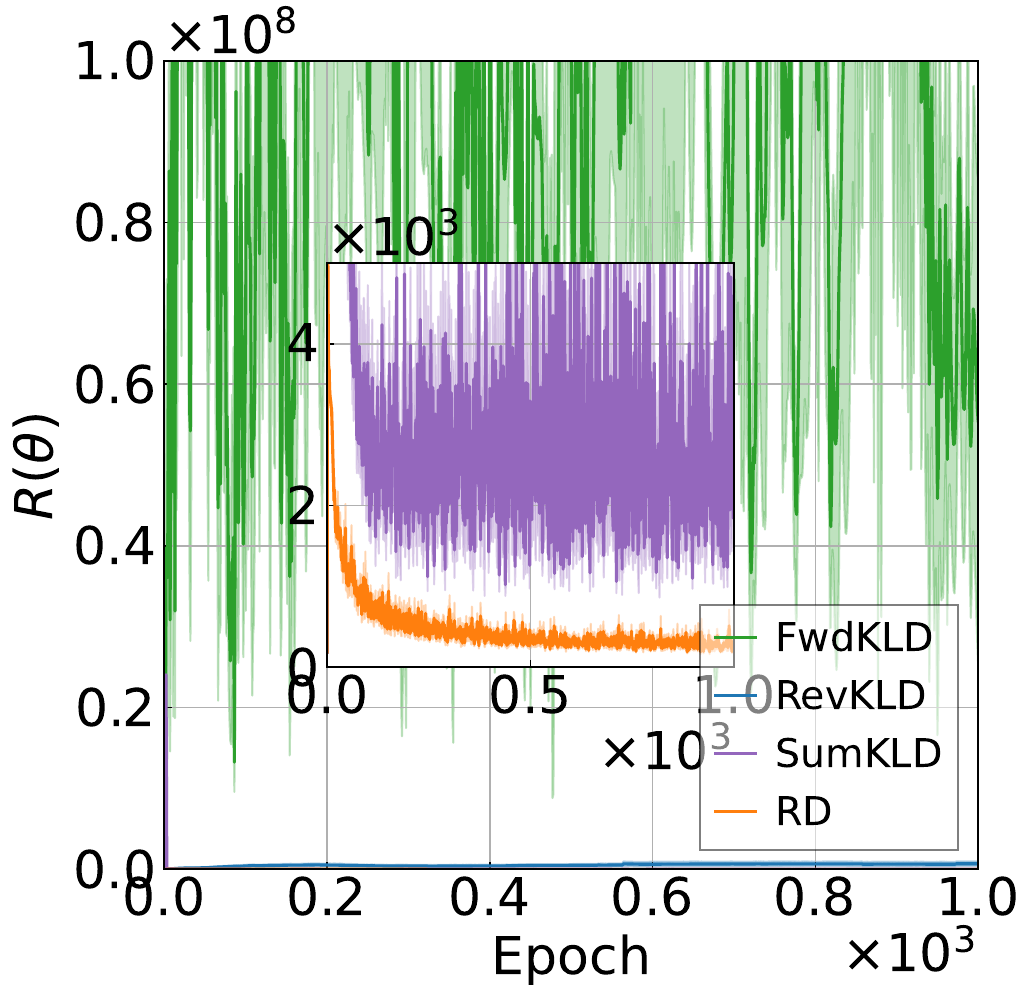}
                \put(0,95){\subfiglabelsize\captiontext*{}}
            \end{overpic}
        \end{minipage}
        \hfill
        \begin{minipage}[t]{0.49\linewidth}
            \centering
            \phantomcaption
            \label{fig:g1_energy_dist}
            \begin{overpic}[width=0.95\textwidth]{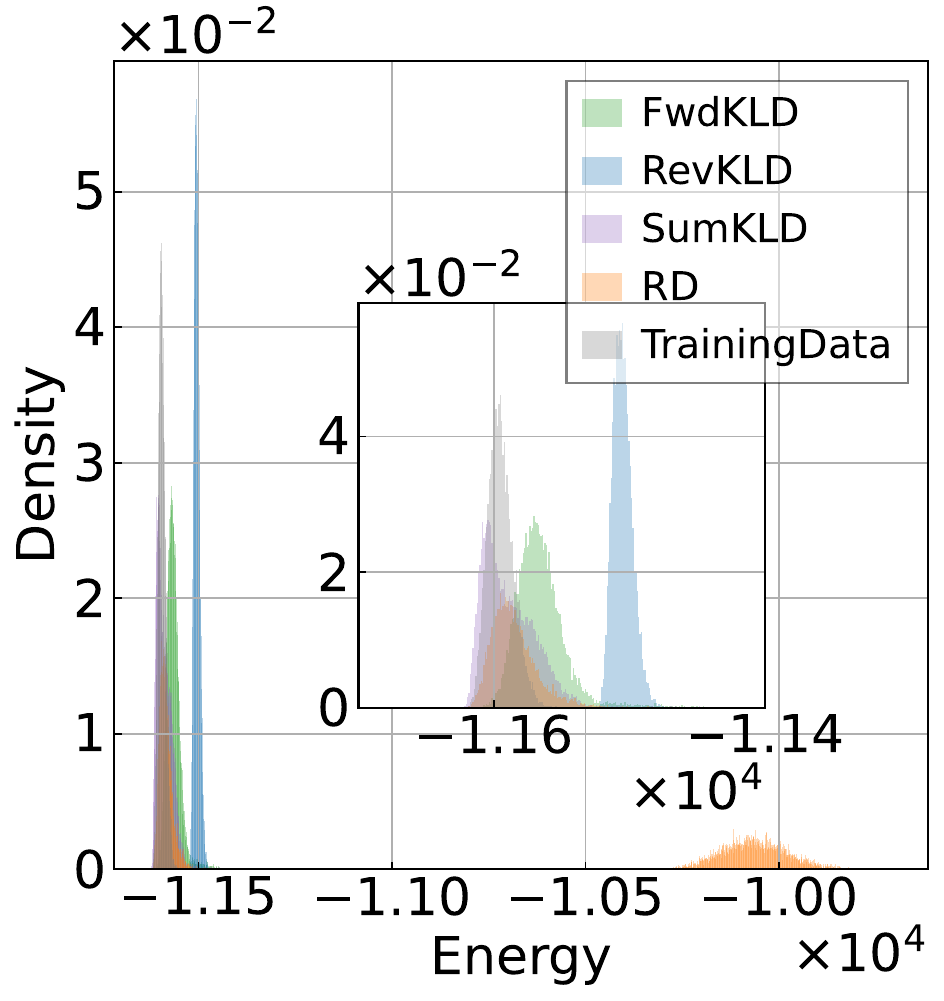}
                \put(0,95){\subfiglabelsize\captiontext*{}}
            \end{overpic}
        \end{minipage}
    \end{subcaptiongroup}
    \caption{\subref{fig:g1_riw} $R(\theta)$ as functions of epochs during the training process of RBMs and \subref{fig:g1_energy_dist} energy distributions of generated samples by each model and those in the training dataset on Gset G1 of the MCP.}
    \label{fig:g1_riw_and_energy_dist}
\end{figure}

\begin{figure}[t]
    \centering
    \begin{subcaptiongroup}
        \begin{minipage}[t]{0.49\linewidth}
            \centering
            \phantomcaption
            \label{fig:g1_pca_fwdkl}
            \begin{overpic}[width=\textwidth]{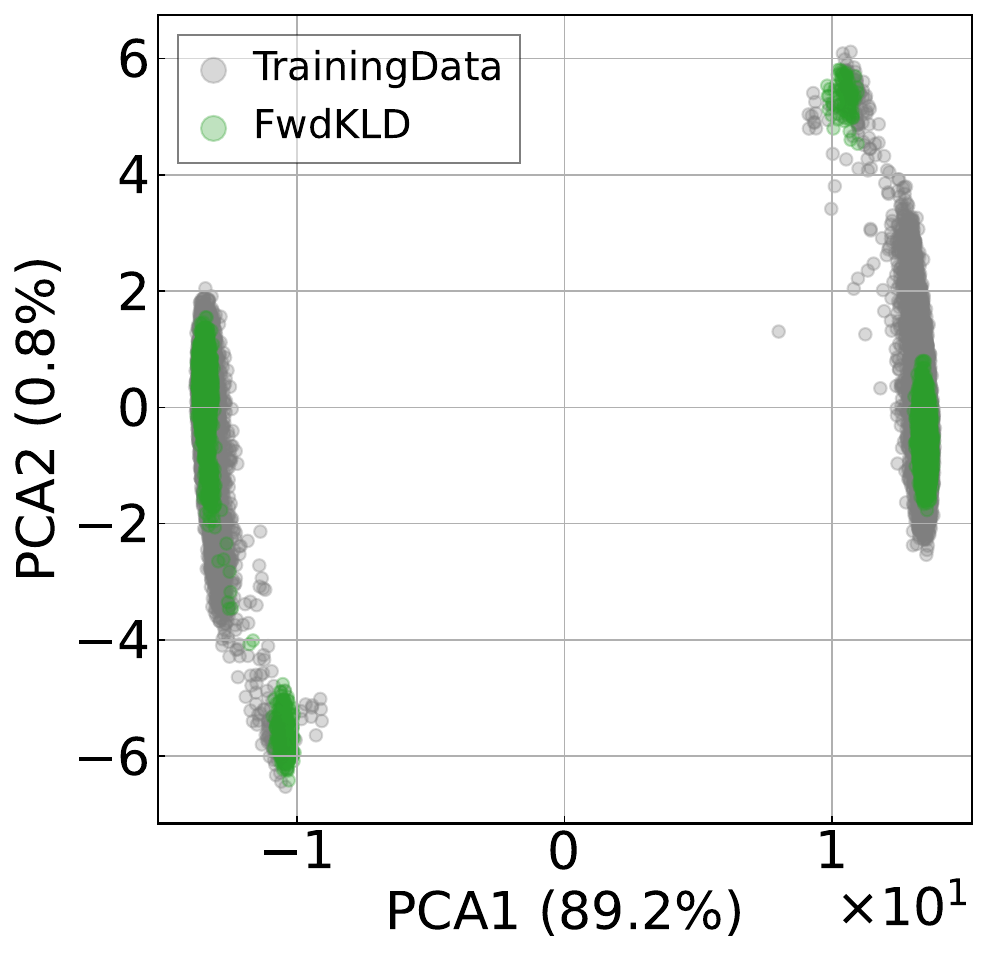}
                \put(0,95){\subfiglabelsize\captiontext*{}}
            \end{overpic}
        \end{minipage}
        \hfill
        \begin{minipage}[t]{0.49\linewidth}
            \centering
            \phantomcaption
            \label{fig:g1_pca_revkl}
            \begin{overpic}[width=\textwidth]{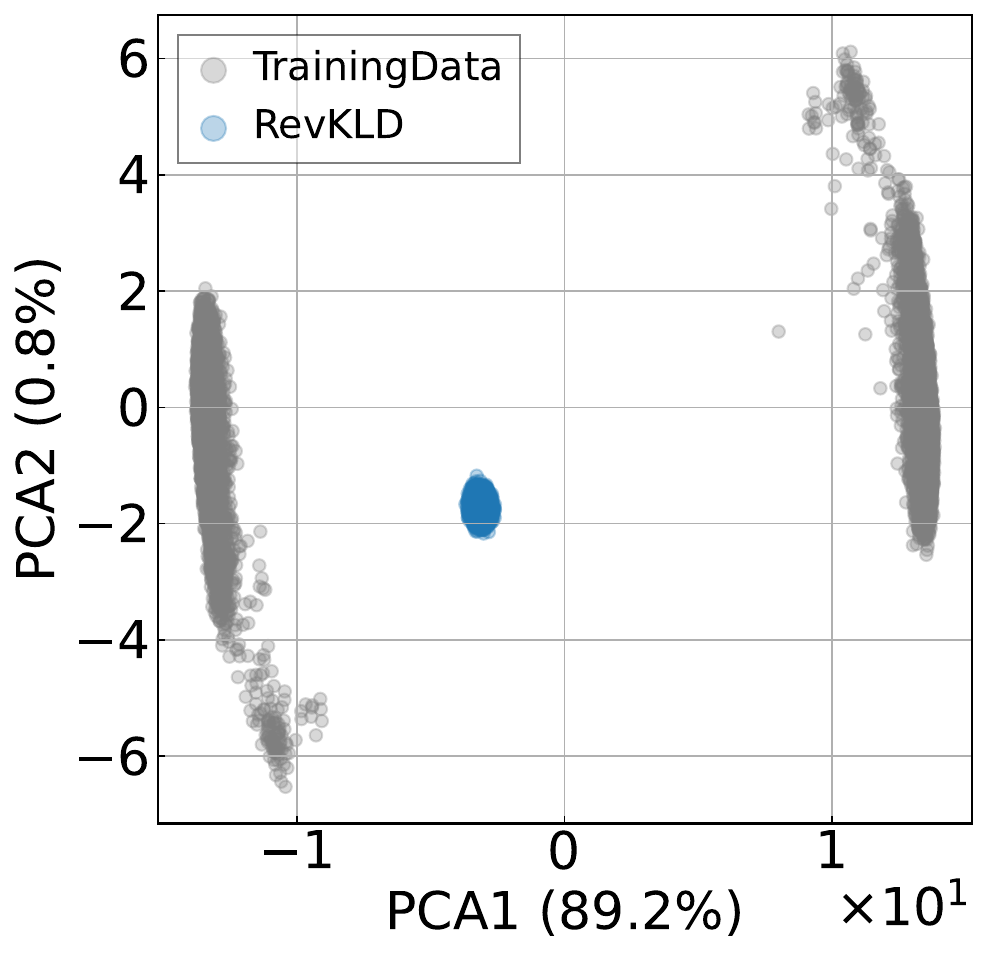}
                \put(0,95){\subfiglabelsize\captiontext*{}}
            \end{overpic}
        \end{minipage}
        \hfill
        \begin{minipage}[t]{0.49\linewidth}
            \centering
            \phantomcaption
            \label{fig:g1_pca_fwdrevkl}
            \begin{overpic}[width=\textwidth]{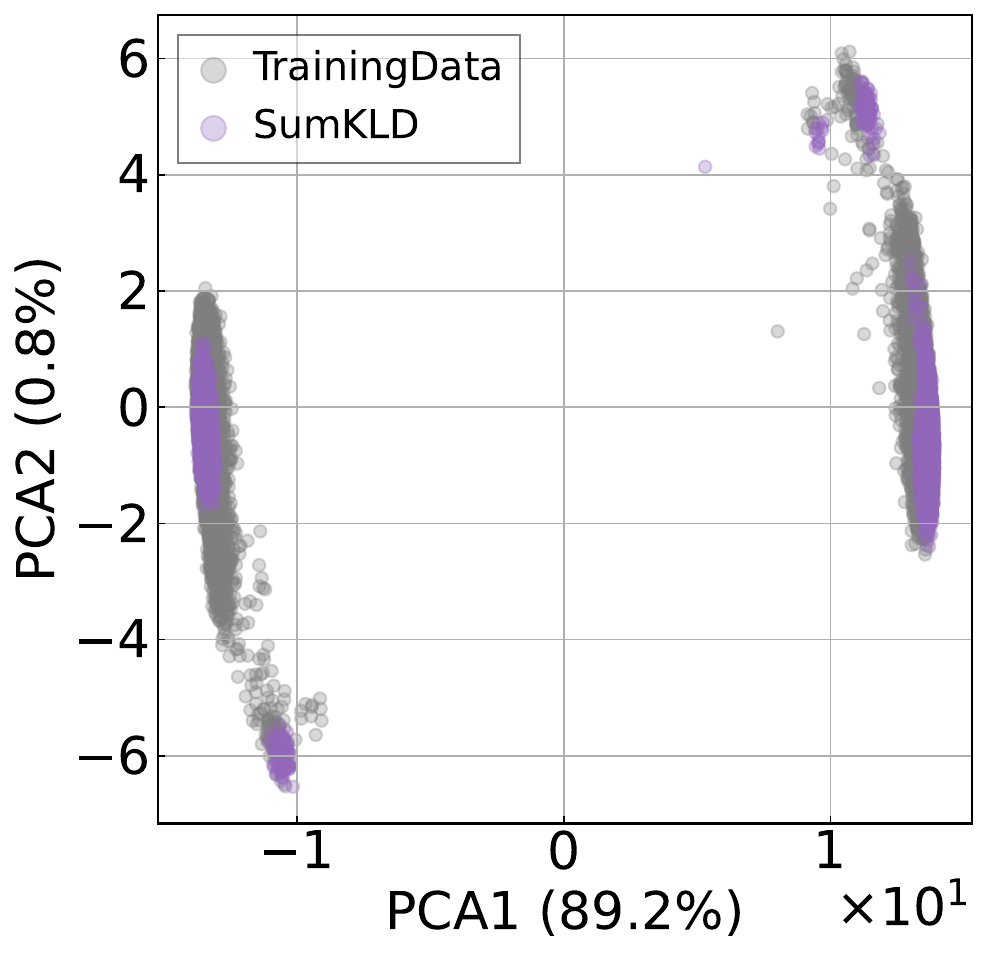}
                \put(0,95){\subfiglabelsize\captiontext*{}}
            \end{overpic}
        \end{minipage}
        \hfill
        \begin{minipage}[t]{0.49\linewidth}
            \centering
            \phantomcaption
            \label{fig:g1_pca_rd}
            \begin{overpic}[width=\textwidth]{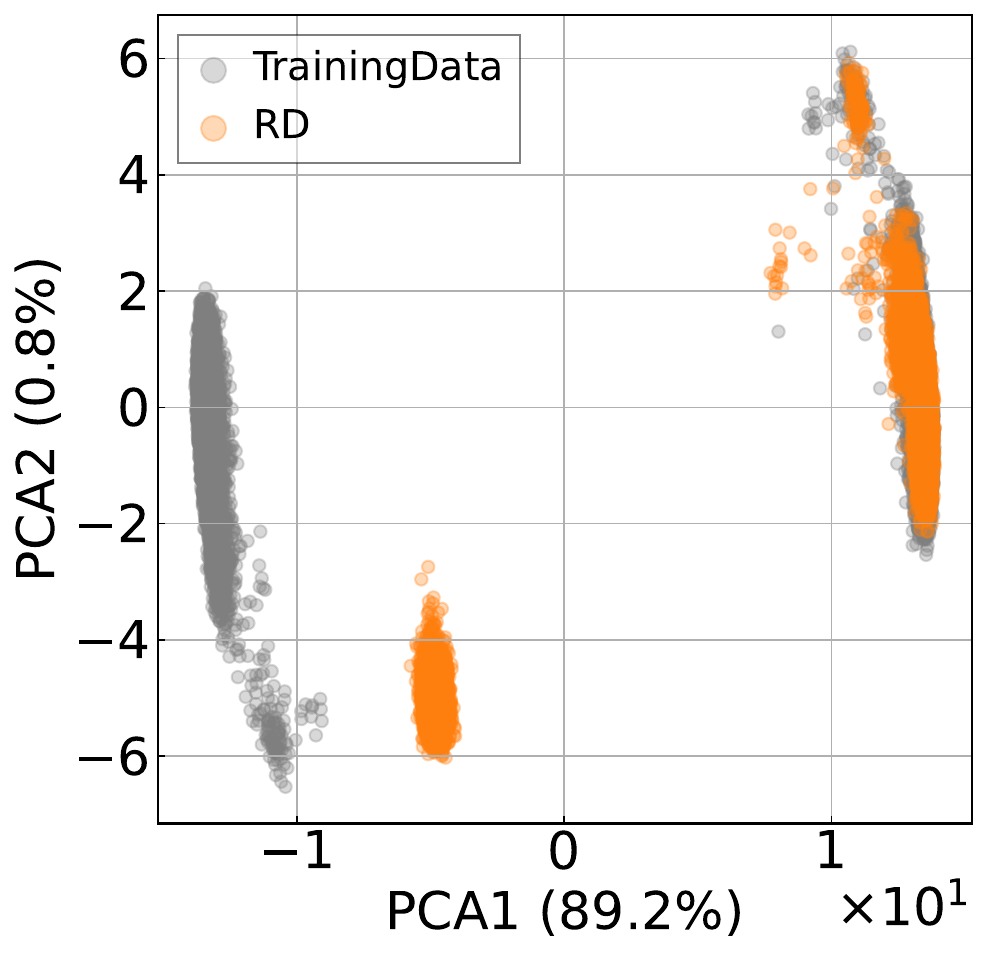}
                \put(0,95){\subfiglabelsize\captiontext*{}}
            \end{overpic}
        \end{minipage}
    \end{subcaptiongroup}
    \caption{\subref{fig:g1_pca_fwdkl}--\subref{fig:g1_pca_rd} Two dimensional PCA mapping of generated samples by each method for the G1 of MCP.}
    \label{fig:g1_pca}
\end{figure}

Figure~\ref{fig:g1_riw} shows that RD learning outperformed the other learning methods at $1000$ epochs in terms of $R(\theta)$.
Moreover, RD learning exhibits significantly high learning stability for $R(\theta)$. 
In contrast, summation KLD learning shows fluctuating learning dynamics in  $R(\theta)$, which may be due to confusion between the dominance of forward KLD or reverse KLD, caused by the simplistic additive approach.

\begin{figure}[t]
    \centering
    \includegraphics[width=\linewidth]{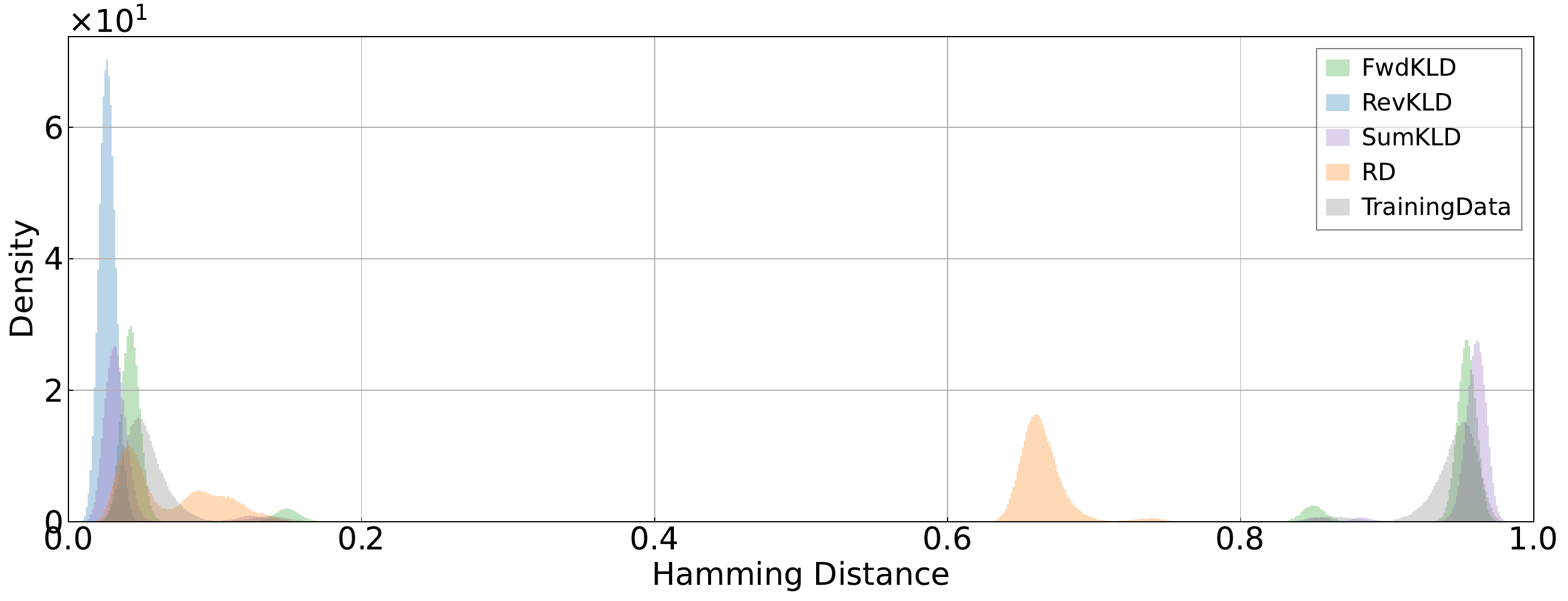}
    \caption{Empirical distribution of the hamming distance of the generated samples by each method and those in the training dataset.}
    \label{fig:g1_hamming_dist}
\end{figure}

While Table~\ref{tab:wasserstein_metric} shows RD learning is not necessarily better than other learning methods, 
Figures~\ref{fig:g1_pca_rd} and \ref{fig:g1_hamming_dist} indicate that RD learning effectively captures two peaks in the PCA space, although it does not recover one of the peaks as accurately as forward KLD and summation KLD learning. 
Note that forward KLD and summation KLD learning suffer from lower energy regression performance compared to RD learning concerning $R(\theta)$.
As a result, while RD learning may not consistently appear superior in every measurement, its overall performance in terms of mode diversity, energy regression, and learning stability makes it a more reliable and effective approach.
Therefore, we consider RD learning to be a promising tool for practical applications, providing a relevant balance of relatively high stability and performance.

\section{Conclusion}\label{sec:conclusion}
This study addresses practical challenges in approximating high-dimensional distributions, specifically underfitting in forward KLD learning and mode collapse in reverse KLD learning, by proposing RD learning.
This approach effectively leverages the strengths of both forward and reverse KLD learning while maintaining low computational cost.
Moreover, this approach is guaranteed to increase the acceptance probability in the Metropolis-Hastings algorithms, leading to enhanced MCMC simulations.
Numerical experiments demonstrate that RD learning overcomes the limitations of complex distribution modeling and scales efficiently with increasing dimensionality.
Furthermore, RD learning exhibits higher learning stability for the regression performance than summation KLD learning.
Consequently, compared to traditional forward and reverse KLD learning and its simple summation, RD learning significantly improves performance and has the potential to enhance generative models with tractable target energy functions.
This advancement opens new avenues for large-scale simulations previously considered infeasible and offers deeper insights into high-dimensional energy landscapes.

\begin{acknowledgments}
This work was supported by JST Grant Number JPMJPF2221 and JPSJ Grant-in-Aid for Scientific Research Number 23H01095. 
Additionally, Yuma Ichikawa was supported by the WINGS-FMSP program at the University of Tokyo.
\end{acknowledgments}

\medskip

\bibliography{ref}

\begin{thebibliography}{38}
\providecommand{\natexlab}[1]{#1}
\providecommand{\url}[1]{\texttt{#1}}
\expandafter\ifx\csname urlstyle\endcsname\relax
  \providecommand{\doi}[1]{doi: #1}\else
  \providecommand{\doi}{doi: \begingroup \urlstyle{rm}\Url}\fi

\bibitem[Baumg{\"a}rtner et~al.(2012)Baumg{\"a}rtner, Burkitt, Ceperley, De~Raedt, Ferrenberg, Heermann, Herrmann, Landau, Levesque, von~der Linden, et~al.]{baumgartner2012monte}
Artur Baumg{\"a}rtner, AN~Burkitt, DM~Ceperley, H~De~Raedt, AM~Ferrenberg, DW~Heermann, HJ~Herrmann, DP~Landau, D~Levesque, W~von~der Linden, et~al.
\newblock \emph{The Monte Carlo method in condensed matter physics}, volume~71.
\newblock Springer Science \& Business Media, 2012.

\bibitem[Binder et~al.(1993)Binder, Heermann, Roelofs, Mallinckrodt, and McKay]{binder1993monte}
Kurt Binder, Dieter Heermann, Lyle Roelofs, A~John Mallinckrodt, and Susan McKay.
\newblock Monte carlo simulation in statistical physics.
\newblock \emph{Computers in Physics}, 7\penalty0 (2):\penalty0 156--157, 1993.

\bibitem[Chen et~al.(2018)Chen, Tan, and Ferguson]{chen2018collective}
Wei Chen, Aik~Rui Tan, and Andrew~L Ferguson.
\newblock Collective variable discovery and enhanced sampling using autoencoders: Innovations in network architecture and error function design.
\newblock \emph{The Journal of chemical physics}, 149\penalty0 (7), 2018.

\bibitem[Ciarella et~al.(2023)Ciarella, Trinquier, Weigt, and Zamponi]{ciarella2023machine}
Simone Ciarella, Jeanne Trinquier, Martin Weigt, and Francesco Zamponi.
\newblock Machine-learning-assisted monte carlo fails at sampling computationally hard problems.
\newblock \emph{Machine Learning: Science and Technology}, 4\penalty0 (1):\penalty0 010501, 2023.

\bibitem[Dabiri et~al.(2020)Dabiri, Malekmohammadi, Sheikholeslami, and Tamura]{dabiri2020replica}
Keivan Dabiri, Mehrdad Malekmohammadi, Ali Sheikholeslami, and Hirotaka Tamura.
\newblock Replica exchange mcmc hardware with automatic temperature selection and parallel trial.
\newblock \emph{IEEE Transactions on Parallel and Distributed Systems}, 31\penalty0 (7):\penalty0 1681--1692, 2020.
\newblock \doi{10.1109/TPDS.2020.2972359}.

\bibitem[Fan et~al.(2021)Fan, Shen, Nussinov, Liu, Sun, and Liu]{fan2021finding}
Changjun Fan, Mutian Shen, Zohar Nussinov, Zhong Liu, Yizhou Sun, and Yang-Yu Liu.
\newblock Finding spin glass ground states through deep reinforcement learning.
\newblock \emph{arXiv preprint arXiv:2109.14411}, 2021.

\bibitem[Fernau et~al.(2019)Fernau, Golovach, Sagot, et~al.]{fernau2019algorithmic}
Henning Fernau, Petr Golovach, Marie-France Sagot, et~al.
\newblock Algorithmic enumeration: Output-sensitive, input-sensitive, parameterized, approximative (dagstuhl seminar 18421).
\newblock In \emph{Dagstuhl Reports}, volume~8. Schloss Dagstuhl-Leibniz-Zentrum fuer Informatik, 2019.

\bibitem[Foreman-Mackey et~al.(2013)Foreman-Mackey, Hogg, Lang, and Goodman]{foreman2013emcee}
Daniel Foreman-Mackey, David~W Hogg, Dustin Lang, and Jonathan Goodman.
\newblock emcee: the mcmc hammer.
\newblock \emph{Publications of the Astronomical Society of the Pacific}, 125\penalty0 (925):\penalty0 306, 2013.

\bibitem[Gelman et~al.(1995)Gelman, Carlin, Stern, and Rubin]{gelman1995bayesian}
Andrew Gelman, John~B Carlin, Hal~S Stern, and Donald~B Rubin.
\newblock \emph{Bayesian data analysis}.
\newblock Chapman and Hall/CRC, 1995.

\bibitem[Gu and Zhang(2022)]{gu2022thermodynamics}
Jing Gu and Kai Zhang.
\newblock Thermodynamics of the ising model encoded in restricted boltzmann machines.
\newblock \emph{Entropy}, 24\penalty0 (12):\penalty0 1701, 2022.

\bibitem[Hanaka et~al.(2023)Hanaka, Kiyomi, Kobayashi, Kobayashi, Kurita, and Otachi]{hanaka2023framework}
Tesshu Hanaka, Masashi Kiyomi, Yasuaki Kobayashi, Yusuke Kobayashi, Kazuhiro Kurita, and Yota Otachi.
\newblock A framework to design approximation algorithms for finding diverse solutions in combinatorial problems.
\newblock In \emph{Proceedings of the AAAI Conference on Artificial Intelligence}, volume~37, pages 3968--3976, 2023.

\bibitem[Hibat-Allah et~al.(2021)Hibat-Allah, Inack, Wiersema, Melko, and Carrasquilla]{hibat2021variational}
Mohamed Hibat-Allah, Estelle~M Inack, Roeland Wiersema, Roger~G Melko, and Juan Carrasquilla.
\newblock Variational neural annealing.
\newblock \emph{Nature Machine Intelligence}, 3\penalty0 (11):\penalty0 952--961, 2021.

\bibitem[Hinton(2002)]{hinton2002training}
Geoffrey~E Hinton.
\newblock Training products of experts by minimizing contrastive divergence.
\newblock \emph{Neural computation}, 14\penalty0 (8):\penalty0 1771--1800, 2002.

\bibitem[Hjelm et~al.(2014)Hjelm, Calhoun, Salakhutdinov, Allen, Adali, and Plis]{hjelm2014restricted}
R~Devon Hjelm, Vince~D Calhoun, Ruslan Salakhutdinov, Elena~A Allen, Tulay Adali, and Sergey~M Plis.
\newblock Restricted boltzmann machines for neuroimaging: an application in identifying intrinsic networks.
\newblock \emph{NeuroImage}, 96:\penalty0 245--260, 2014.

\bibitem[Huang and Wang(2017)]{huang2017accelerated}
Li~Huang and Lei Wang.
\newblock Accelerated {{Monte Carlo}} simulations with restricted {{Boltzmann}} machines.
\newblock \emph{Physical Review B}, 95\penalty0 (3):\penalty0 035105, January 2017.
\newblock \doi{10.1103/PhysRevB.95.035105}.

\bibitem[Hukushima and Nemoto(1996)]{Hukushima_1996_J.Phys.Soc.Jpn.}
Koji Hukushima and Koji Nemoto.
\newblock Exchange {{Monte Carlo Method}} and {{Application}} to {{Spin Glass Simulations}}.
\newblock \emph{Journal of the Physical Society of Japan}, 65\penalty0 (6):\penalty0 1604--1608, June 1996.
\newblock ISSN 0031-9015.
\newblock \doi{10.1143/JPSJ.65.1604}.

\bibitem[Hyvärinen(2007)]{HYVARINEN20072499}
Aapo Hyvärinen.
\newblock Some extensions of score matching.
\newblock \emph{Computational Statistics \& Data Analysis}, 51\penalty0 (5):\penalty0 2499--2512, 2007.
\newblock ISSN 0167-9473.
\newblock \doi{https://doi.org/10.1016/j.csda.2006.09.003}.
\newblock URL \url{https://www.sciencedirect.com/science/article/pii/S0167947306003264}.

\bibitem[Inack et~al.(2022)Inack, Morawetz, and Melko]{inack2022neural}
Estelle~M Inack, Stewart Morawetz, and Roger~G Melko.
\newblock Neural annealing and visualization of autoregressive neural networks in the newman--moore model.
\newblock \emph{Condensed Matter}, 7\penalty0 (2):\penalty0 38, 2022.

\bibitem[Khajenezhad et~al.(2020)Khajenezhad, Madani, and Beigy]{khajenezhad2020masked}
Ahmad Khajenezhad, Hatef Madani, and Hamid Beigy.
\newblock Masked autoencoder for distribution estimation on small structured data sets.
\newblock \emph{IEEE Transactions on Neural Networks and Learning Systems}, 32\penalty0 (11):\penalty0 4997--5007, 2020.

\bibitem[Kingma and Ba(2014)]{Kingma_2014_arXiv.org}
Diederik~P. Kingma and Jimmy Ba.
\newblock Adam: {{A Method}} for {{Stochastic Optimization}}.
\newblock \emph{arXiv.org}, December 2014.

\bibitem[Le~Roux and Bengio(2008)]{le2008representational}
Nicolas Le~Roux and Yoshua Bengio.
\newblock Representational power of restricted boltzmann machines and deep belief networks.
\newblock \emph{Neural computation}, 20\penalty0 (6):\penalty0 1631--1649, 2008.

\bibitem[Liu et~al.(2017)Liu, Qi, Meng, and Fu]{liu2017self}
Junwei Liu, Yang Qi, Zi~Yang Meng, and Liang Fu.
\newblock Self-learning monte carlo method.
\newblock \emph{Physical Review B}, 95\penalty0 (4):\penalty0 041101, 2017.

\bibitem[Martin et~al.(2011)Martin, Quinn, and Park]{martin2011mcmcpack}
Andrew~D Martin, Kevin~M Quinn, and Jong~Hee Park.
\newblock Mcmcpack: Markov chain monte carlo in r.
\newblock 2011.

\bibitem[McNaughton et~al.(2020)McNaughton, Milo{\v{s}}evi{\'c}, Perali, and Pilati]{mcnaughton2020boosting}
B~McNaughton, MV~Milo{\v{s}}evi{\'c}, A~Perali, and S~Pilati.
\newblock Boosting monte carlo simulations of spin glasses using autoregressive neural networks.
\newblock \emph{Physical Review E}, 101\penalty0 (5):\penalty0 053312, 2020.

\bibitem[Melko et~al.(2019)Melko, Carleo, Carrasquilla, and Cirac]{melko2019restricted}
Roger~G Melko, Giuseppe Carleo, Juan Carrasquilla, and J~Ignacio Cirac.
\newblock Restricted boltzmann machines in quantum physics.
\newblock \emph{Nature Physics}, 15\penalty0 (9):\penalty0 887--892, 2019.

\bibitem[Midgley et~al.(2022)Midgley, Stimper, Simm, Sch{\"o}lkopf, and Hern{\'a}ndez-Lobato]{midgley2022flow}
Laurence~Illing Midgley, Vincent Stimper, Gregor~NC Simm, Bernhard Sch{\"o}lkopf, and Jos{\'e}~Miguel Hern{\'a}ndez-Lobato.
\newblock Flow annealed importance sampling bootstrap.
\newblock \emph{arXiv preprint arXiv:2208.01893}, 2022.

\bibitem[Puente and Eremin(2020)]{puente2020convolutional}
Daniel~Alcalde Puente and Ilya~M Eremin.
\newblock Convolutional restricted boltzmann machine aided monte carlo: an application to ising and kitaev models.
\newblock \emph{Physical Review B}, 102\penalty0 (19):\penalty0 195148, 2020.

\bibitem[Ribeiro et~al.(2018)Ribeiro, Bravo, Wang, and Tiwary]{ribeiro2018reweighted}
Jo{\~a}o Marcelo~Lamim Ribeiro, Pablo Bravo, Yihang Wang, and Pratyush Tiwary.
\newblock Reweighted autoencoded variational bayes for enhanced sampling (rave).
\newblock \emph{The Journal of chemical physics}, 149\penalty0 (7), 2018.

\bibitem[Smolensky et~al.(1986)]{smolensky1986information}
Paul Smolensky et~al.
\newblock Information processing in dynamical systems: Foundations of harmony theory.
\newblock 1986.

\bibitem[Stimper et~al.(2022)Stimper, Sch{\"o}lkopf, and Hern{\'a}ndez-Lobato]{stimper2022resampling}
Vincent Stimper, Bernhard Sch{\"o}lkopf, and Jos{\'e}~Miguel Hern{\'a}ndez-Lobato.
\newblock Resampling base distributions of normalizing flows.
\newblock In \emph{International Conference on Artificial Intelligence and Statistics}, pages 4915--4936. PMLR, 2022.

\bibitem[Tieleman(2008)]{Tieleman_2008_Proc.25thInt.Conf.Mach.Learn.}
Tijmen Tieleman.
\newblock Training restricted {{Boltzmann}} machines using approximations to the likelihood gradient.
\newblock In \emph{Proceedings of the 25th International Conference on {{Machine}} Learning}, {{ICML}} '08, pages 1064--1071, {New York, NY, USA}, July 2008. {Association for Computing Machinery}.
\newblock ISBN 978-1-60558-205-4.
\newblock \doi{10.1145/1390156.1390290}.

\bibitem[Uria et~al.(2016)Uria, C{\^o}t{\'e}, Gregor, Murray, and Larochelle]{uria2016neural}
Benigno Uria, Marc-Alexandre C{\^o}t{\'e}, Karol Gregor, Iain Murray, and Hugo Larochelle.
\newblock Neural autoregressive distribution estimation.
\newblock \emph{The Journal of Machine Learning Research}, 17\penalty0 (1):\penalty0 7184--7220, 2016.

\bibitem[Wirnsberger et~al.(2022)Wirnsberger, Papamakarios, Ibarz, Racani{\`e}re, Ballard, Pritzel, and Blundell]{wirnsberger2022normalizing}
Peter Wirnsberger, George Papamakarios, Borja Ibarz, S{\'e}bastien Racani{\`e}re, Andrew~J Ballard, Alexander Pritzel, and Charles Blundell.
\newblock Normalizing flows for atomic solids.
\newblock \emph{Machine Learning: Science and Technology}, 3\penalty0 (2):\penalty0 025009, 2022.

\bibitem[Wu et~al.(2019)Wu, Wang, and Zhang]{wu2019solving}
Dian Wu, Lei Wang, and Pan Zhang.
\newblock Solving statistical mechanics using variational autoregressive networks.
\newblock \emph{Physical review letters}, 122\penalty0 (8):\penalty0 080602, 2019.

\bibitem[Wu et~al.(2021)Wu, Rossi, and Carleo]{wu2021unbiased}
Dian Wu, Riccardo Rossi, and Giuseppe Carleo.
\newblock Unbiased monte carlo cluster updates with autoregressive neural networks.
\newblock \emph{Physical Review Research}, 3\penalty0 (4):\penalty0 L042024, 2021.

\bibitem[Wu et~al.(2020)Wu, K{\"o}hler, and No{\'e}]{wu2020stochastic}
Hao Wu, Jonas K{\"o}hler, and Frank No{\'e}.
\newblock Stochastic normalizing flows.
\newblock \emph{Advances in Neural Information Processing Systems}, 33:\penalty0 5933--5944, 2020.

\bibitem[Xu et~al.(2017)Xu, Qi, Liu, Fu, and Meng]{xu2017self}
Xiao~Yan Xu, Yang Qi, Junwei Liu, Liang Fu, and Zi~Yang Meng.
\newblock Self-learning quantum monte carlo method in interacting fermion systems.
\newblock \emph{Physical Review B}, 96\penalty0 (4):\penalty0 041119, 2017.

\bibitem[Ye(2003)]{Gset}
Yinyu Ye.
\newblock Gset, 2003.
\newblock URL \url{http://web.stanford.edu/~yyye/yyye/Gset/}.

\end{thebibliography}
\bibliographystyle{plainnat}

\onecolumngrid
\appendix

\section{Proof of Ratio divergence properties}\label{seq:proof_of_ratio_divergence}

\subsection{Symmetric divergence}\label{subsec:proof_of_symmetric_divergence}
In this section, we prove that Eq.~\eqref{eq:ratio-divergence} is a symmetric divergence.
To this end, since it is trivial that Eq.~\eqref{eq:ratio-divergence} is symmetric and non-negative, we need to prove $\mac{L}(\hat{P},P;\B{\theta})\Leftrightarrow \hat{P}=P$.
Eq.~\eqref{eq:ratio-divergence} sums up the square of $\log \frac{\hat{P}(\B{x}')P(\B{x};\B{\theta})}{P(\B{x}';\B{\theta})\hat{P}(\B{x})}$.
Therefore, if $\mathcal{L}(\hat{P},P;\B{\theta}) = 0$, then the following function satisfies for all $\B{x}$ and $\B{x}'$:
\begin{equation}
    \log \frac{\hat{P}(\B{x}')P(\B{x};\B{\theta})}{P(\B{x}';\B{\theta})\hat{P}(\B{x})} = 0
    \Leftrightarrow
    \frac{\hat{P}(\B{x}')}{\hat{P}(\B{x})} = \frac{P(\B{x}';\B{\theta})}{P(\B{x};\B{\theta})}.
\end{equation}
This equation straightforwardly leads to $\hat{P}(\B{x}) = P(\B{x};\B{\theta})$ for all $\B{x}$ since $\hat{P}$ and $P$ are probabilities.
On the other hand, it is trivial that $\hat{P}=P \Rightarrow \mathcal{L}(\hat{P},P;\B{\theta}) = 0$.

\subsection{Guarantee of increasing acceptance probability}\label{subsec:proof_of_inequality_related_with_MH}
We denote $A(\B{x}',\B{x})$ as the acceptance probability in the Metropolis-Hastings algorithm, that is
\begin{equation}
    A(\B{x}',\B{x}) = \min \left(1, \frac{\hat{P}(\B{x}^{\prime}) P(\B{x};\B{\theta})}{P(\B{x}^{\prime};\B{\theta})\hat{P}(\B{x})} \right).
\end{equation}
For any $\B{x}'$ and $\B{x}$, $\hat{P}$ and $P$ satisfy the following inequality:
\begin{align}
    \left(\log \frac{\hat{P}(\B{x}^{\prime}) P(\B{x};\B{\theta})}{P(\B{x}^{\prime};\B{\theta})\hat{P}(\B{x})} \right)^{2}
    \ge \left(\min \left(0, \log \frac{\hat{P}(\B{x}^{\prime}) P(\B{x};\B{\theta})}{P(\B{x}^{\prime};\B{\theta})\hat{P}(\B{x})} \right)\right)^2 
    = \left(\log A(\B{x}',\B{x}) \right)^2.
\end{align}
Taking the expectation with respect to $\hat{P}$ and $P$ over $\B{x}'$ and $\B{x}$ respectively, and using Jensen's inequality, we obtain the following inequality:
\begin{align}
    \mac{L}(\hat{P},P;\B{\theta})
    \ge \mab{E}_{\hat{P}(\B{x}'),P(\B{x};\B{\theta})} \left[ \left( \log A(\B{x}',\B{x}) \right)^2 \right]
    \ge \left( \mab{E}_{\hat{P}(\B{x}'),P(\B{x};\B{\theta})} \left[ \log A(\B{x}',\B{x}) \right] \right)^2.
\end{align}
Here, considering $\mac{L}(\hat{P},P;\B{\theta}) \ge 0$ and $\mab{E}_{\hat{P}(\B{x}'),P(\B{x};\B{\theta})} \left[ \log A(\B{x}',\B{x}) \right] \le 0$, and using Jensen's inequality, we obtain the following inequality:
\begin{align}
    - \sqrt{\mac{L}(\hat{P},P;\B{\theta})}
    \le \mab{E}_{\hat{P}(\B{x}'),P(\B{x};\B{\theta})} \left[ \log A(\B{x}',\B{x}) \right]
    \le \log \mab{E}_{\hat{P}(\B{x}'),P(\B{x};\B{\theta})} \left[ A(\B{x}',\B{x}) \right].
\end{align}
Then, we obtain the following inequality:
\begin{equation}
    \exp \left( - \sqrt{\mac{L}(\hat{P},P;\B{\theta})} \right)
    \le \mab{E}_{\hat{P}(\B{x}'),P(\B{x};\B{\theta})} \left[ A(\B{x}',\B{x}) \right].
\end{equation}
This is identical to inequality~\eqref{eq:inequality_related_with_MH}.

\section{Additional experiment details}\label{sec:additional_experiment_details}

\subsection{How to prepare training dataset using exchange MCMC simulations}\label{subsec:pt_setting}
In all experiments, we ran exchange MCMC simulations for $10^6$ Monte Carlo steps (MCS), performing replica exchanges every $1$ MCS and recording samples every $10$ MCS, resulting in $10^5$ recorded samples.
After discarding the first $10^4$ recorded samples as a burn-in period, we adopted the next $16{,}384$ samples as the training dataset and the last $1{,}024$ samples as the validation dataset.
For each target model, we set the number of replicas, the minimum temperatures, and the maximum temperature as shown in Table~\ref{tab:pt_parameter}.
We set the intermediate temperatures using the following formula to ensure the temperature ratio is constant:
\begin{equation}
    \beta_i = \beta_1 \exp \left( \frac{i-1}{N-1} \log \frac{\beta_{N_\beta}}{\beta_1} \right),
\end{equation}
where $N_\beta$ is the number of replicas, $\{\beta_i\}_{i=1}^{N_\beta}$ is the temperatures.
\begin{table}[ht]
\caption{Exchange MCMC simulation settings.}
\label{tab:pt_parameter}
\vskip 0.15in
\begin{center}
\begin{small}
\begin{sc}
\begin{tabular}{lccc}
\hline
Target Model & $N_\beta$ & $\beta_1$ & $\beta_{N_\beta}$ \\
\hline
Ising & $4$ & $0.25$ & $0.5$ \\
SK  & $8$ & $0.5$ & $2.0$  \\
MIS & 40 & $0.1$ & $2.0$ \\
Gset G1 & $16$ & $0.25$ & $1.0$ \\
Gset G6 & $16$ & $0.25$ & $2.0$ \\
Gset G14 & $32$ & $0.25$ & $4.0$ \\
Gset G18 & $32$ & $0.25$ & $2.5$ \\
\hline
\end{tabular}
\end{sc}
\end{small}
\end{center}
\vskip -0.1in
\end{table}

\subsection{Wasserstein metric}\label{subsec:wasserstein}
Wasserstein metric of the $L1$ norm between probabilities $P(x)$ and $Q(x)$ on $x \in \mathbb{R}$ is written as follows:
\begin{equation}
    D_W(P,Q) = \int |F_P(x) - F_Q(x)| \mathrm{d}x,
\end{equation}
where $F_P$ and $F_Q$ is the cumulative distribution function of $P$ and $Q$, respectively.
In all experiments, we used SciPy version 1.7.0 to calculate this metric.

\section{Experimental result of Gset on distinct graphs}\label{sec:gset_additional_result}
\begin{figure}
    \centering
    \begin{minipage}[t]{0.49\textwidth}
        \centering
        \begin{subcaptiongroup}
            \begin{minipage}[t]{0.49\textwidth}
                \centering
                \phantomcaption
                \label{fig:g6_riw}
                \begin{overpic}[width=\textwidth]{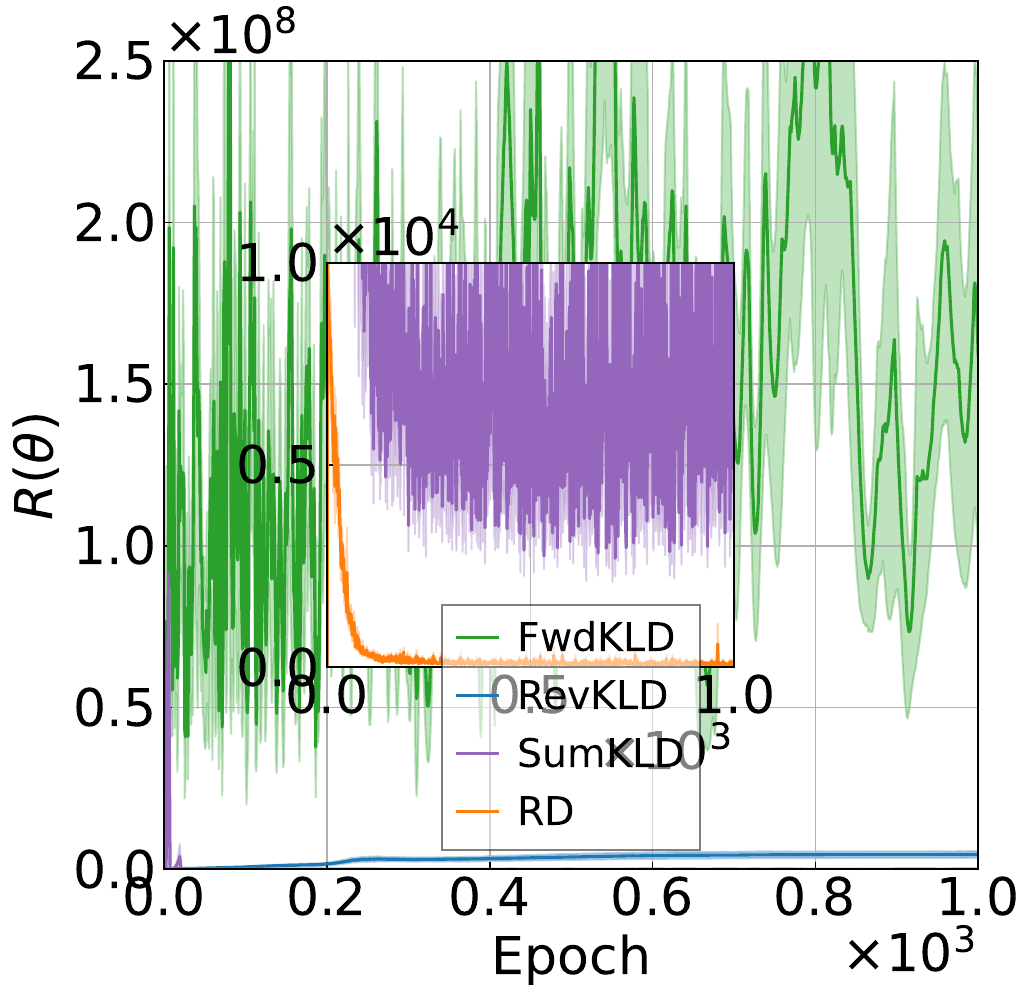}
                    \put(0,95){\subfiglabelsize\captiontext*{}}
                \end{overpic}
            \end{minipage}
            \hfill
            \begin{minipage}[t]{0.49\textwidth}
                \centering
                \phantomcaption
                \label{fig:g6_energy_dist}
                \begin{overpic}[width=\textwidth]{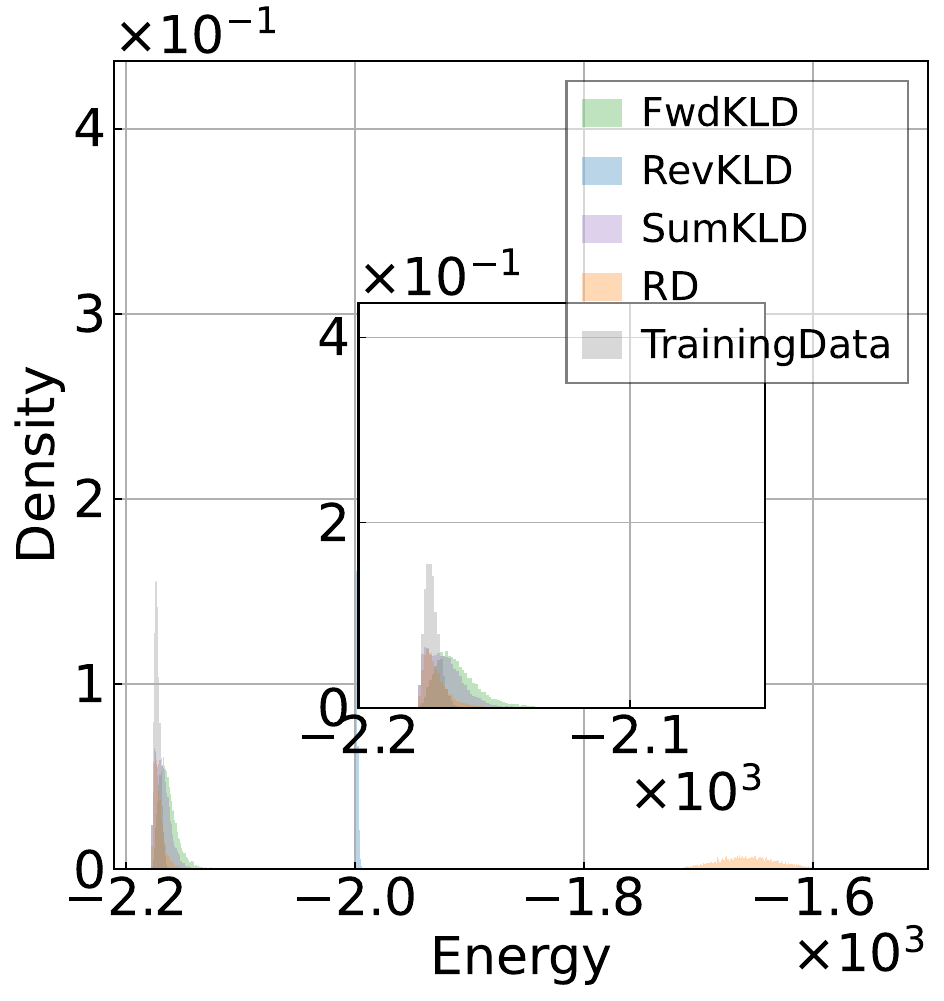}
                    \put(0,95){\subfiglabelsize\captiontext*{}}
                \end{overpic}
            \end{minipage}
        \end{subcaptiongroup}
        \caption{\subref{fig:g6_riw} $R(\theta)$ as functions of epochs during the training process of RBMs and \subref{fig:g6_energy_dist} energy distributions of generated samples by each model and those in the training dataset on Gset G6 of the MCP.}
        \label{fig:g6_riw_and_energy_dist}
    \end{minipage}
    \hfill
    \begin{minipage}[t]{0.49\textwidth}
        \centering
        \begin{subcaptiongroup}
            \begin{minipage}[t]{0.49\textwidth}
                \centering
                \phantomcaption
                \label{fig:g6_pca_fwdkl}
                \begin{overpic}[width=\textwidth]{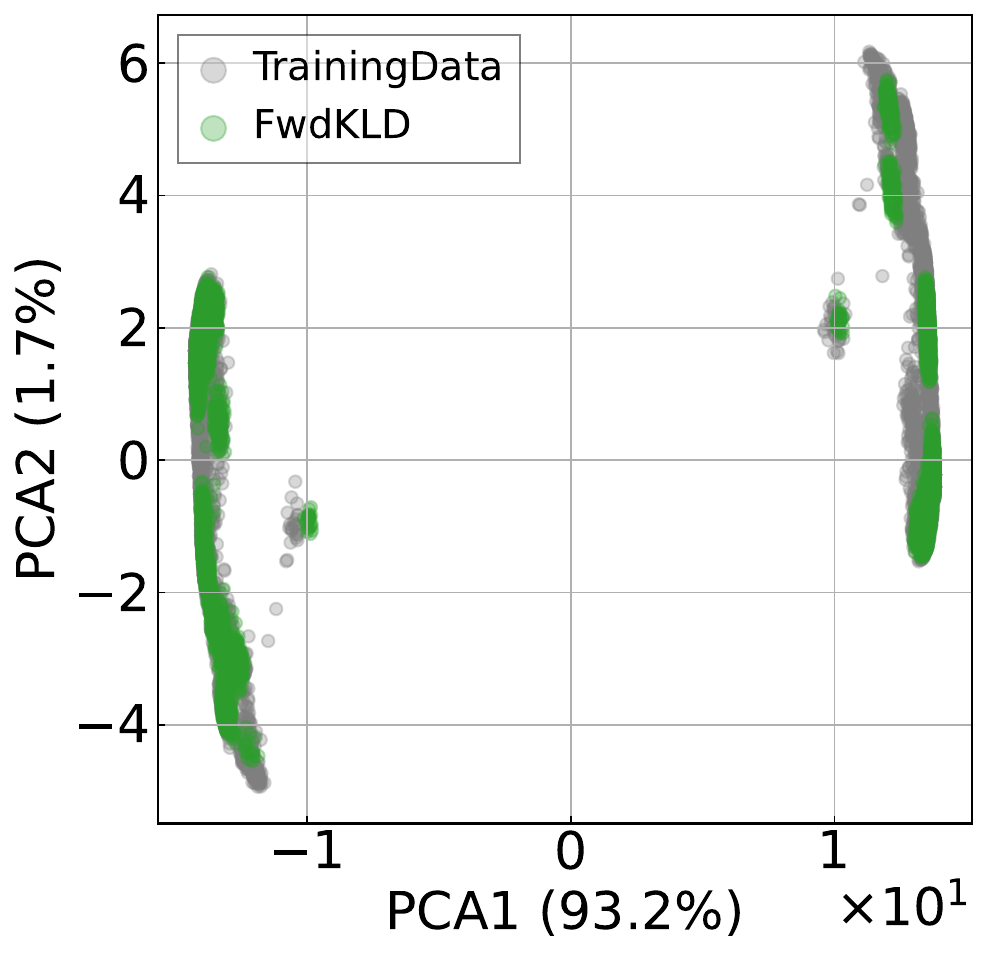}
                    \put(0,95){\subfiglabelsize\captiontext*{}}
                \end{overpic}
            \end{minipage}
            \hfill
            \begin{minipage}[t]{0.49\textwidth}
                \centering
                \phantomcaption
                \label{fig:g6_pca_revkl}
                \begin{overpic}[width=\textwidth]{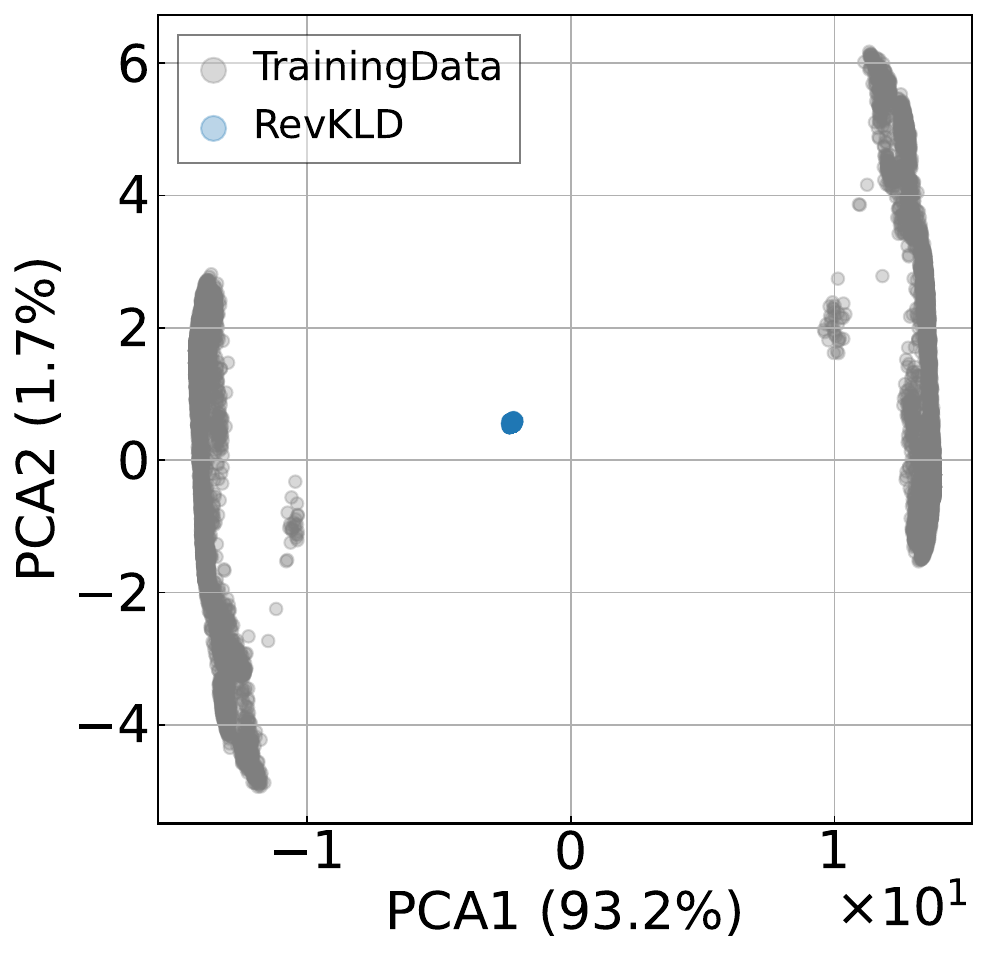}
                    \put(0,95){\subfiglabelsize\captiontext*{}}
                \end{overpic}
            \end{minipage}
            \hfill
            \begin{minipage}[t]{0.49\textwidth}
                \centering
                \phantomcaption
                \label{fig:g6_pca_fwdrevkl}
                \begin{overpic}[width=\textwidth]{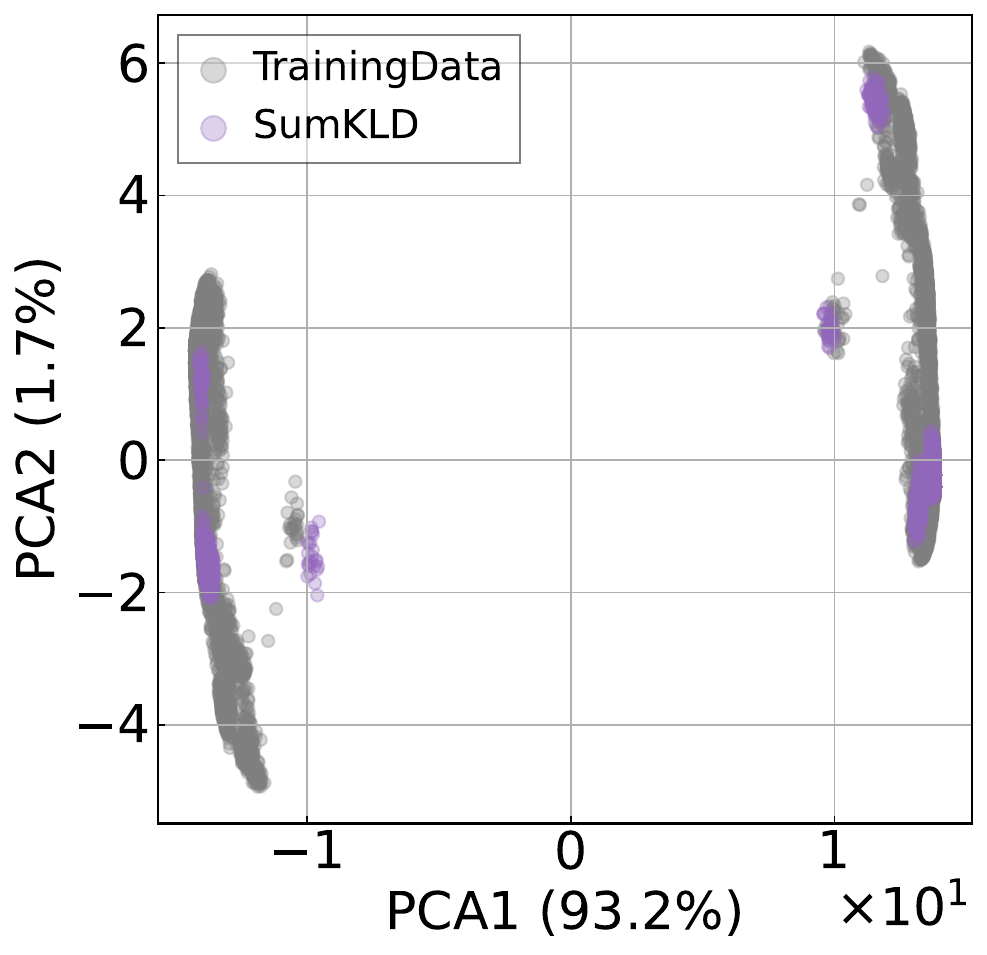}
                    \put(0,95){\subfiglabelsize\captiontext*{}}
                \end{overpic}
            \end{minipage}
            \hfill
            \begin{minipage}[t]{0.49\textwidth}
                \centering
                \phantomcaption
                \label{fig:g6_pca_rd}
                \begin{overpic}[width=\textwidth]{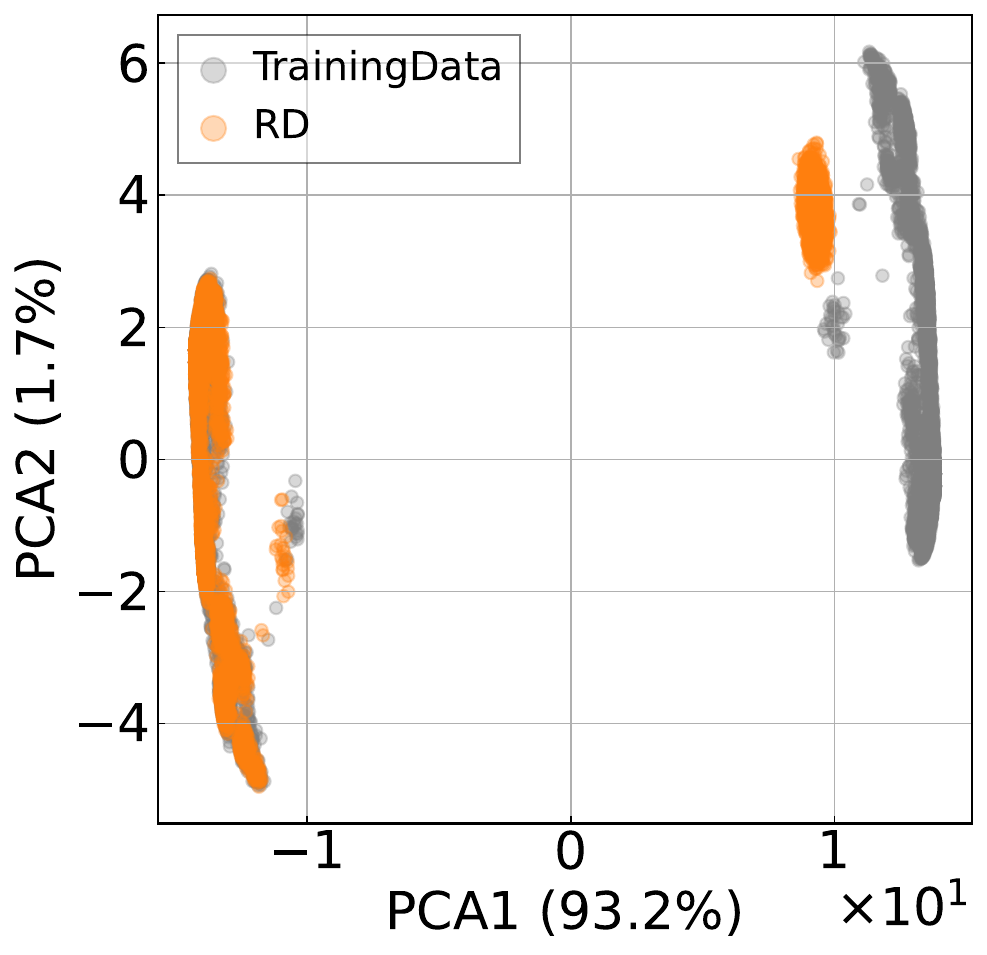}
                    \put(0,95){\subfiglabelsize\captiontext*{}}
                \end{overpic}
            \end{minipage}
        \end{subcaptiongroup}
        \caption{\subref{fig:g6_pca_fwdkl}--\subref{fig:g6_pca_rd} Two dimensional PCA mapping of generated samples by each method for the G6 of MCP.}
        \label{fig:g6_pca}
    \end{minipage}
    \begin{minipage}[t]{0.6\textwidth}
        \centering
        \includegraphics[width=\textwidth]{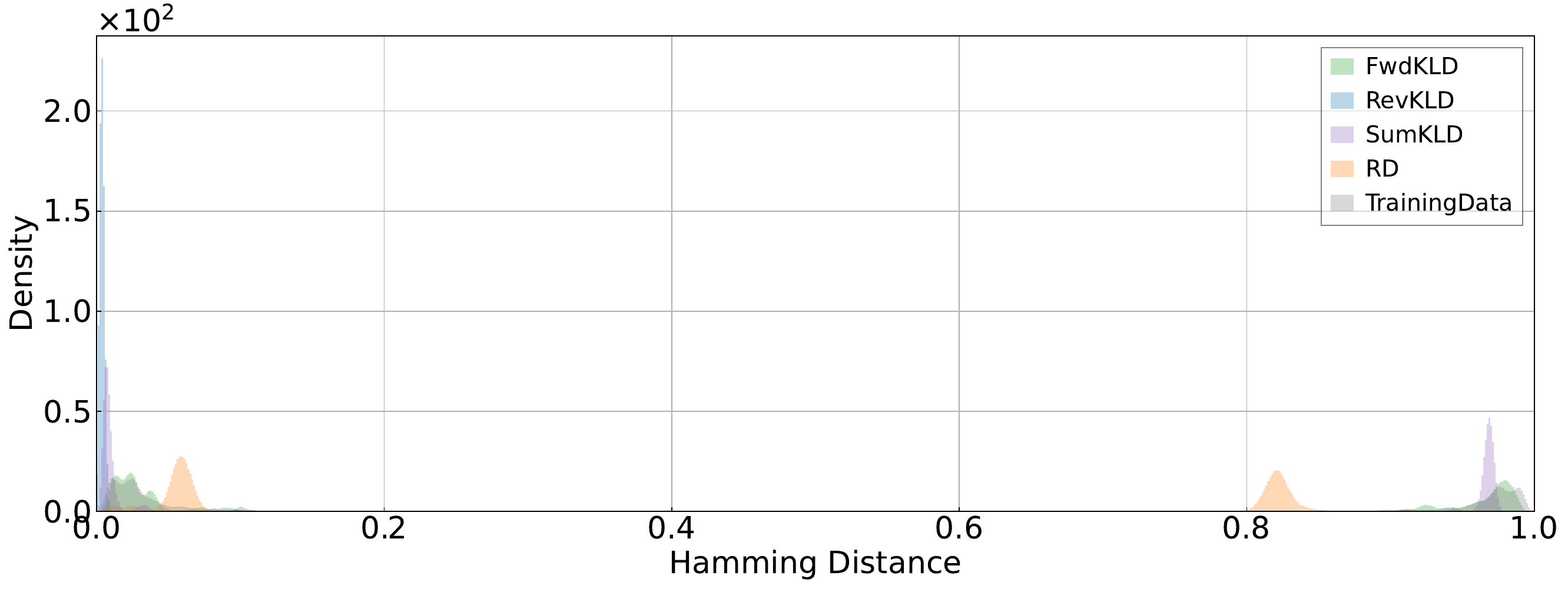}
        \caption{Empirical distribution of the hamming distance of the generated samples by each method and those in the training dataset.}
        \label{fig:g6_hamming_dist}
    \end{minipage}
\end{figure}

\begin{figure}
    \centering
    \begin{minipage}[t]{0.49\textwidth}
        \centering
        \begin{subcaptiongroup}
            \begin{minipage}[t]{0.49\textwidth}
                \centering
                \phantomcaption
                \label{fig:g14_riw}
                \begin{overpic}[width=\textwidth]{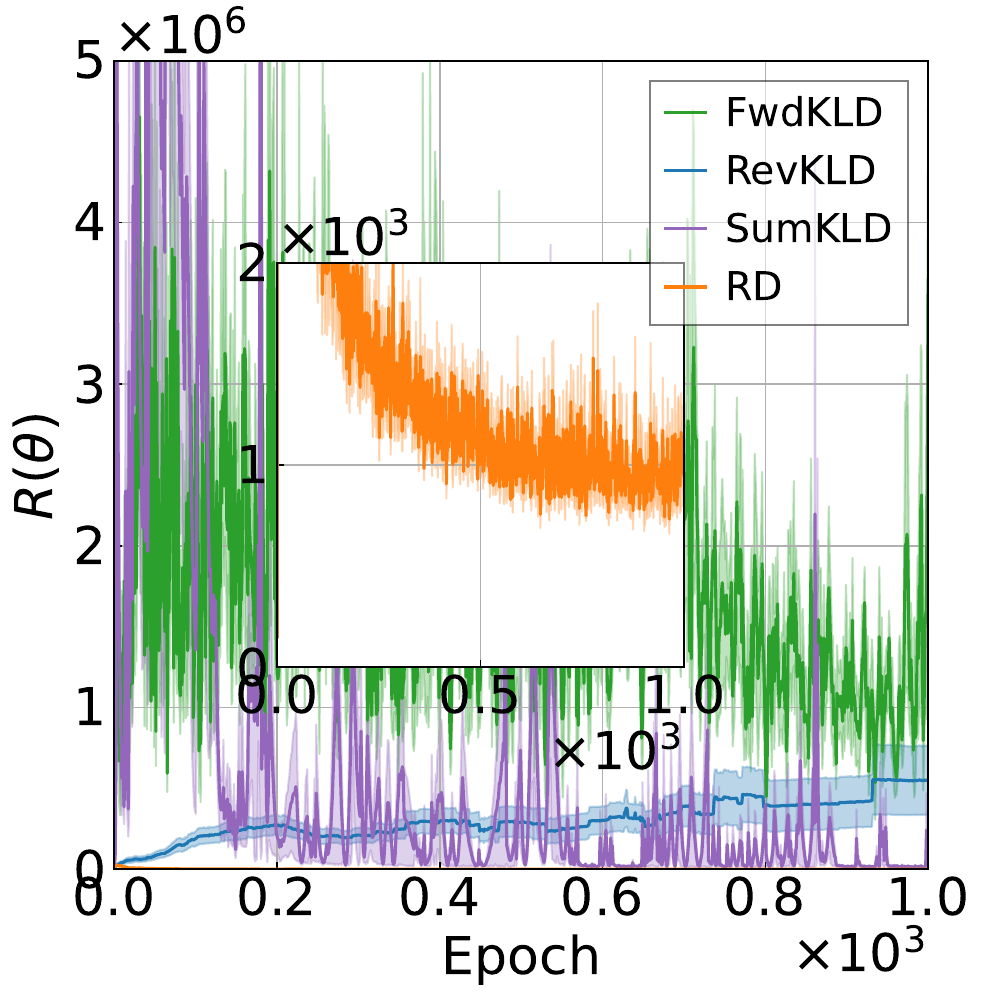}
                    \put(0,95){\subfiglabelsize\captiontext*{}}
                \end{overpic}
            \end{minipage}
            \hfill
            \begin{minipage}[t]{0.49\textwidth}
                \centering
                \phantomcaption
                \label{fig:g14_energy_dist}
                \begin{overpic}[width=\textwidth]{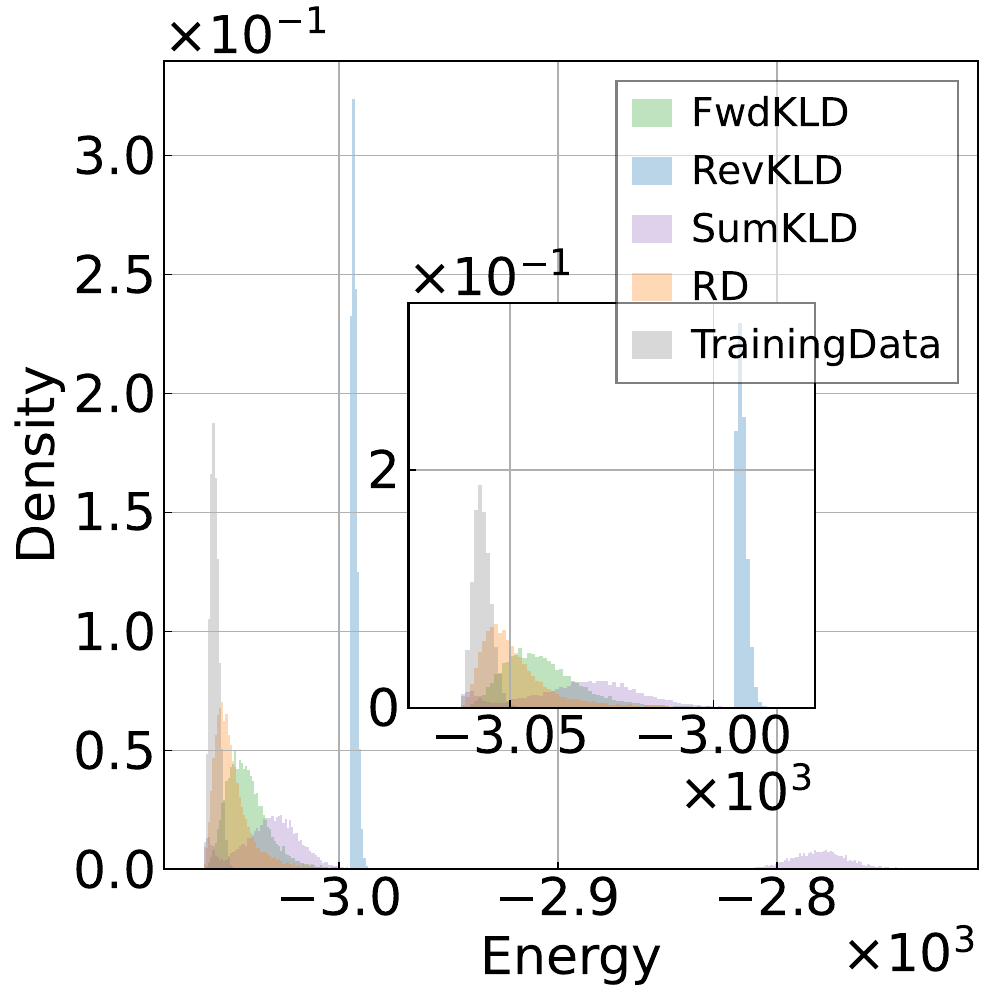}
                    \put(0,95){\subfiglabelsize\captiontext*{}}
                \end{overpic}
            \end{minipage}
        \end{subcaptiongroup}
        \caption{\subref{fig:g14_riw} $R(\theta)$ as functions of epochs during the training process of RBMs and \subref{fig:g14_energy_dist} energy distributions of generated samples by each model and those in the training dataset on Gset G14 of the MCP.}
        \label{fig:g14_riw_and_energy_dist}
    \end{minipage}
    \hfill
    \begin{minipage}[t]{0.49\textwidth}
        \centering
        \begin{subcaptiongroup}
            \begin{minipage}[t]{0.49\textwidth}
                \centering
                \phantomcaption
                \label{fig:g14_pca_fwdkl}
                \begin{overpic}[width=\textwidth]{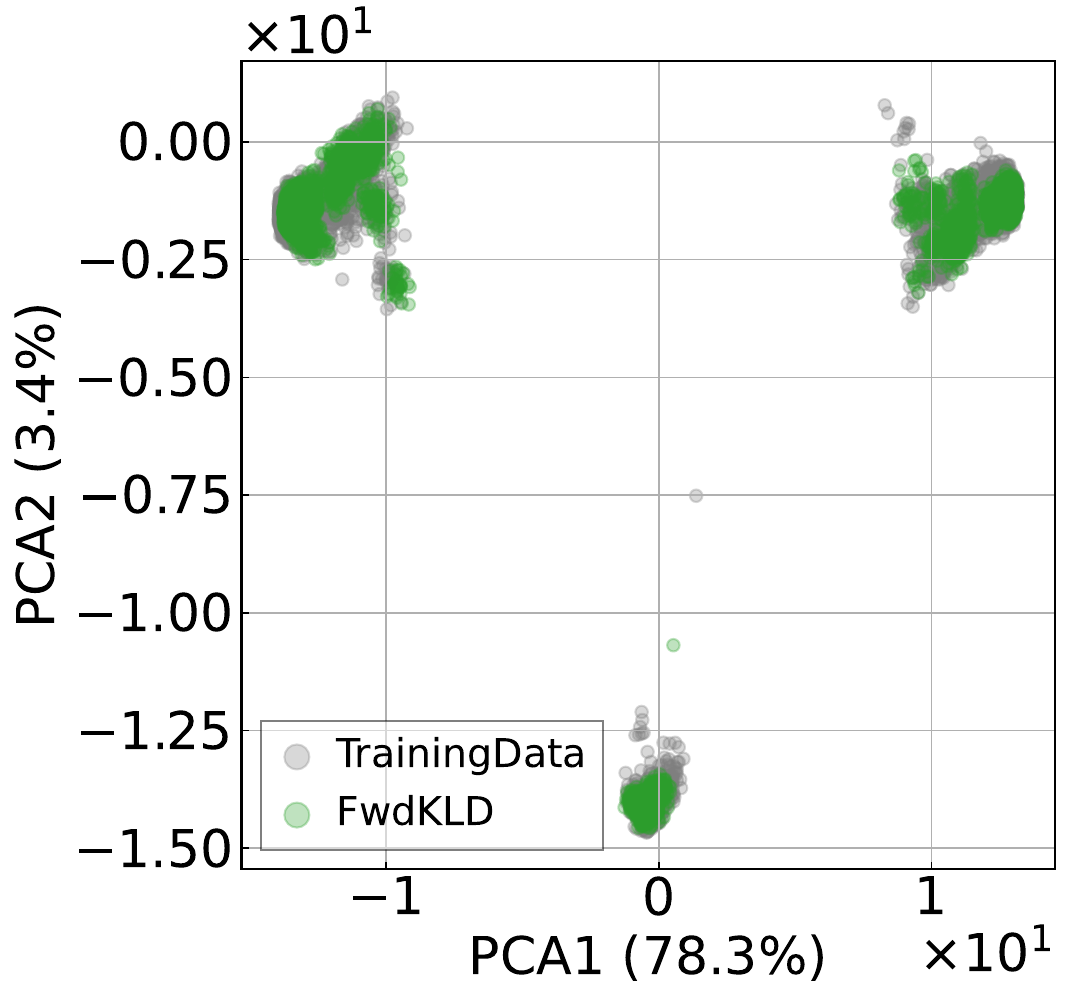}
                    \put(0,95){\subfiglabelsize\captiontext*{}}
                \end{overpic}
            \end{minipage}
            \hfill
            \begin{minipage}[t]{0.49\textwidth}
                \centering
                \phantomcaption
                \label{fig:g14_pca_revkl}
                \begin{overpic}[width=\textwidth]{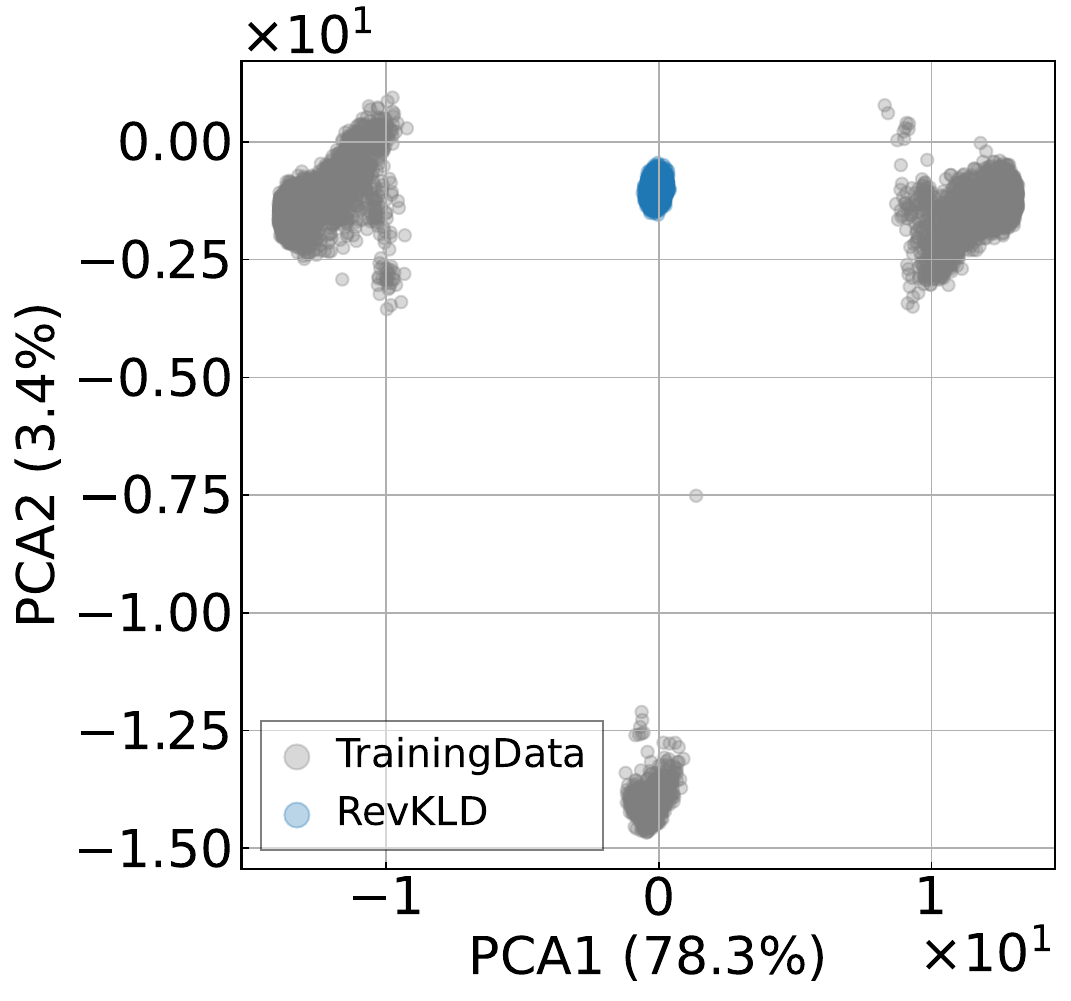}
                    \put(0,95){\subfiglabelsize\captiontext*{}}
                \end{overpic}
            \end{minipage}
            \hfill
            \begin{minipage}[t]{0.49\textwidth}
                \centering
                \phantomcaption
                \label{fig:g14_pca_fwdrevkl}
                \begin{overpic}[width=\textwidth]{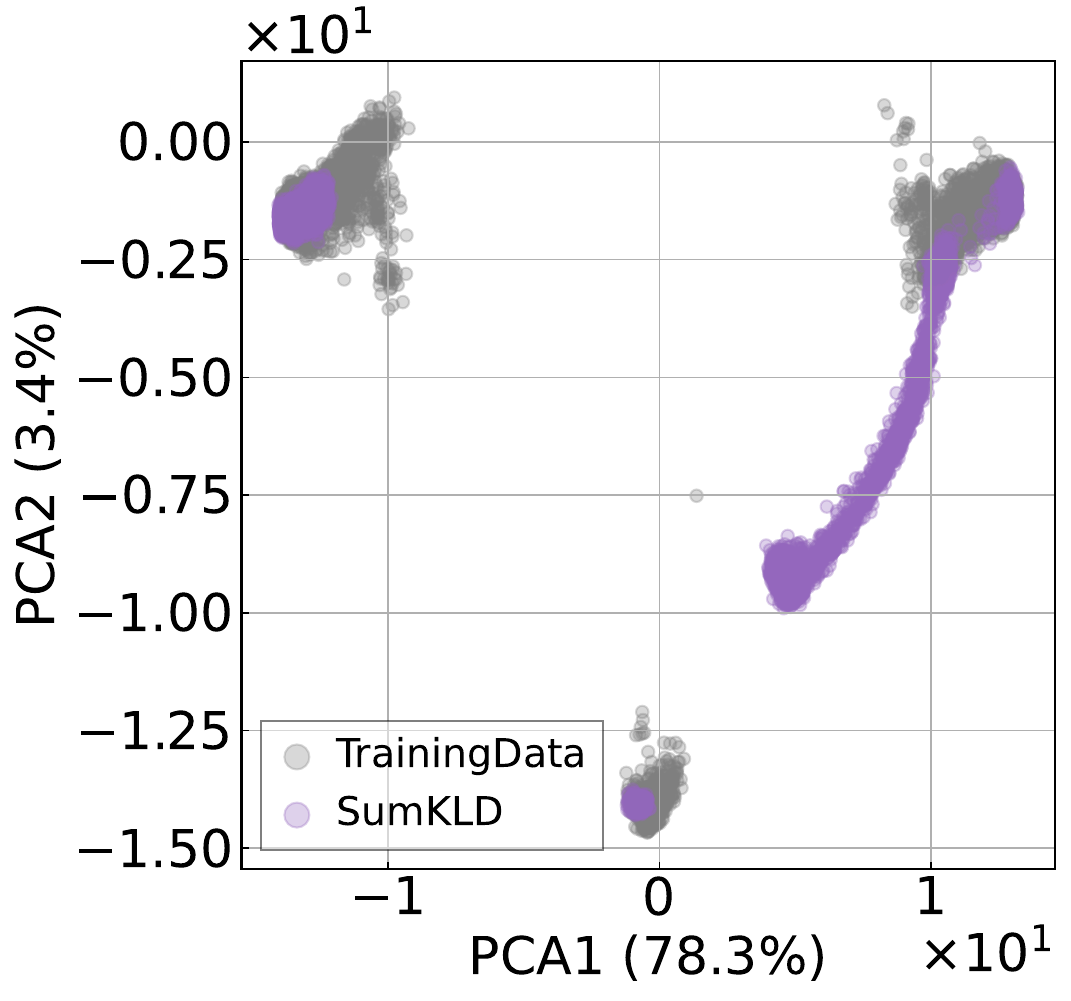}
                    \put(0,95){\subfiglabelsize\captiontext*{}}
                \end{overpic}
            \end{minipage}
            \hfill
            \begin{minipage}[t]{0.49\textwidth}
                \centering
                \phantomcaption
                \label{fig:g14_pca_rd}
                \begin{overpic}[width=\textwidth]{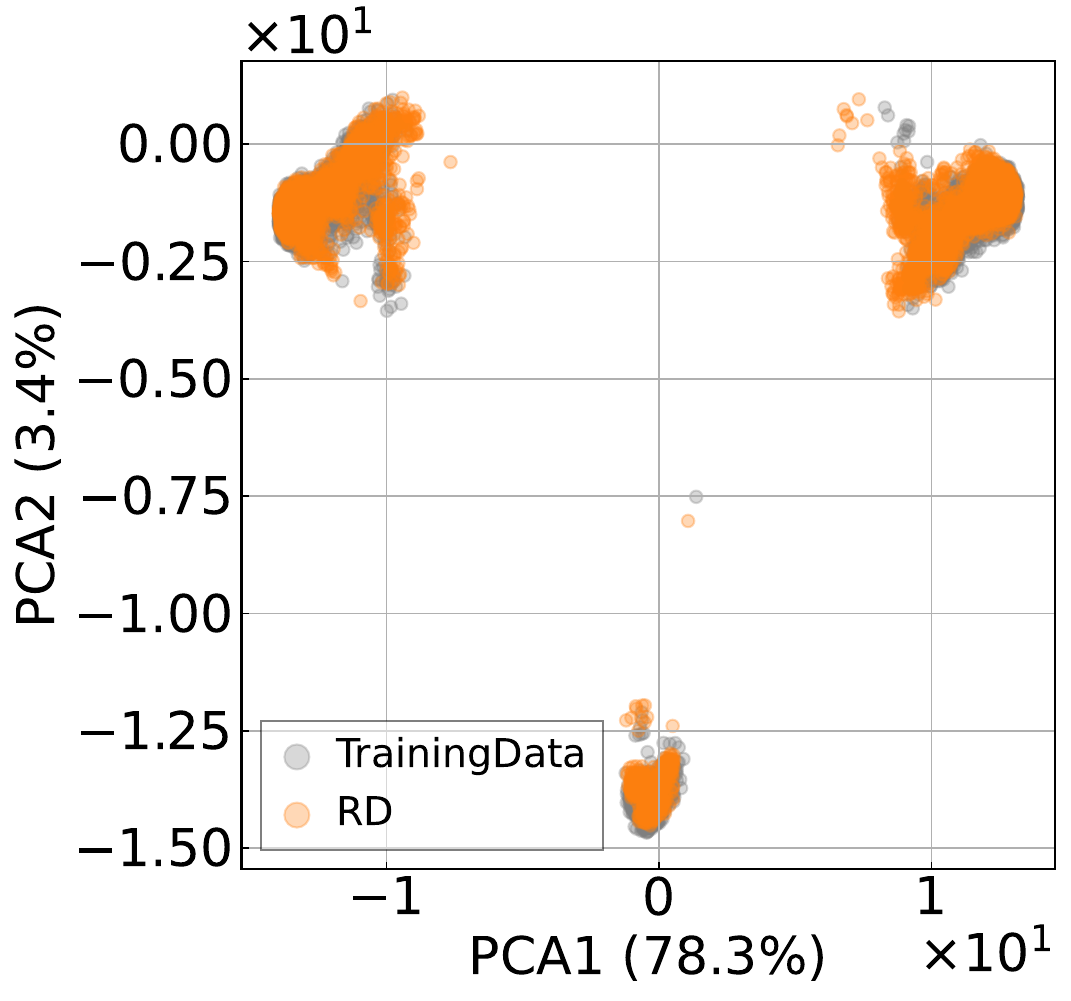}
                    \put(0,95){\subfiglabelsize\captiontext*{}}
                \end{overpic}
            \end{minipage}
        \end{subcaptiongroup}
        \caption{\subref{fig:g14_pca_fwdkl}--\subref{fig:g14_pca_rd} Two dimensional PCA mapping of generated samples by each method for the G14 of MCP.}
        \label{fig:g14_pca}
    \end{minipage}
    \begin{minipage}[t]{0.6\textwidth}
        \centering
        \includegraphics[width=\textwidth]{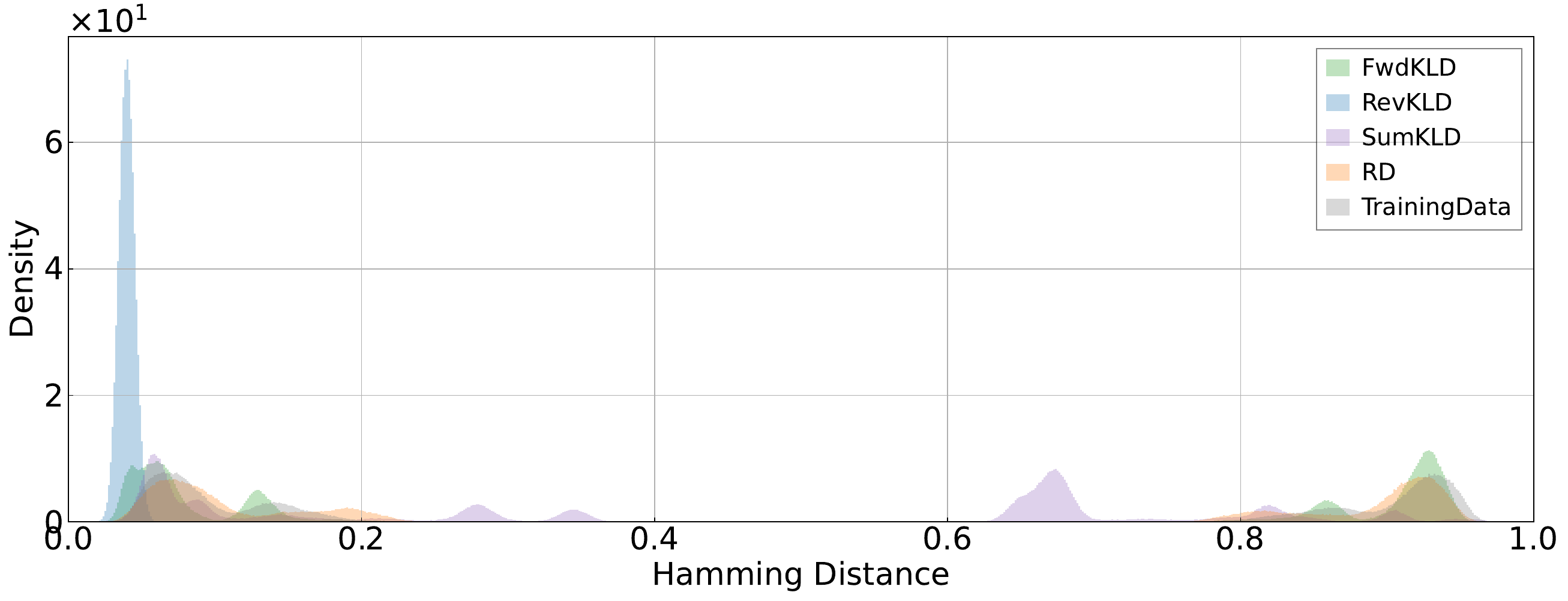}
        \caption{Empirical distribution of the hamming distance of the generated samples by each method and those in the training dataset.}
        \label{fig:g14_hamming_dist}
    \end{minipage}
\end{figure}

\begin{figure}
    \centering
    \begin{minipage}[t]{0.49\textwidth}
        \centering
        \begin{subcaptiongroup}
            \begin{minipage}[t]{0.49\textwidth}
                \centering
                \phantomcaption
                \label{fig:g18_riw}
                \begin{overpic}[width=\textwidth]{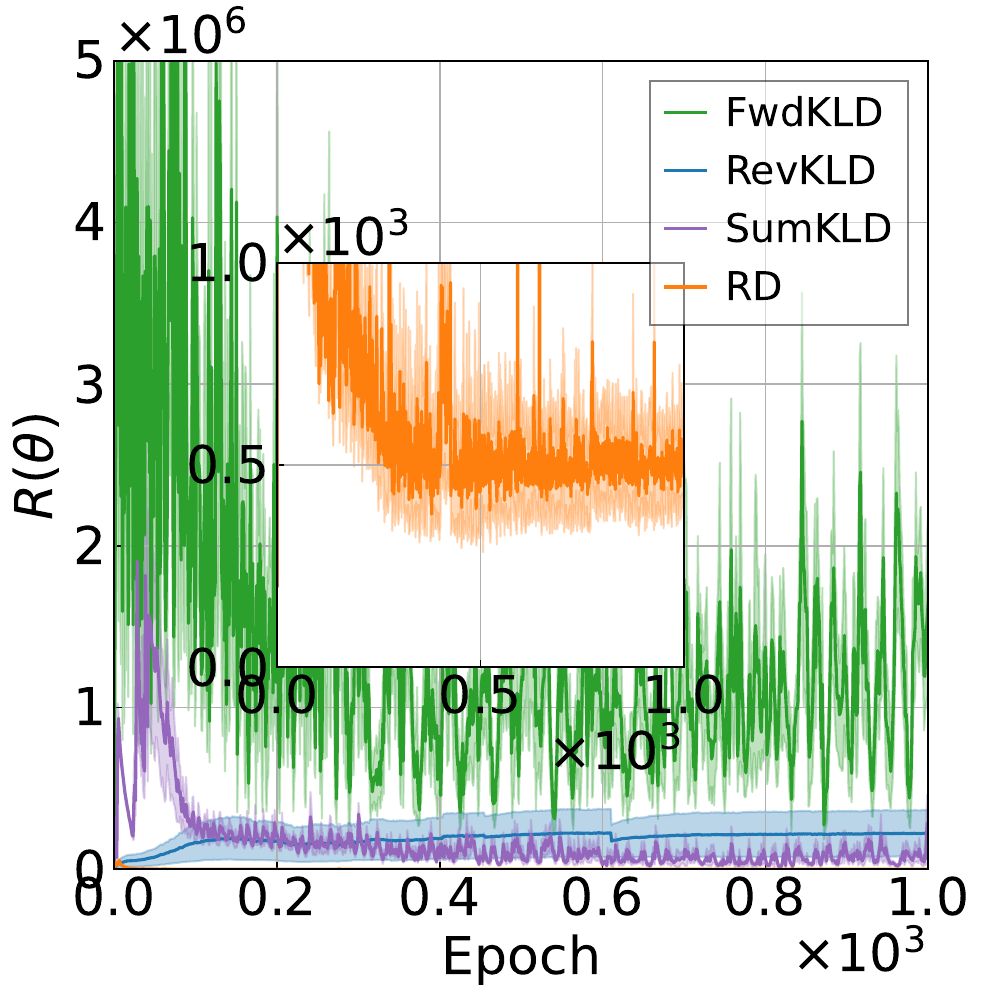}
                    \put(0,95){\subfiglabelsize\captiontext*{}}
                \end{overpic}
            \end{minipage}
            \hfill
            \begin{minipage}[t]{0.49\textwidth}
                \centering
                \phantomcaption
                \label{fig:g18_energy_dist}
                \begin{overpic}[width=\textwidth]{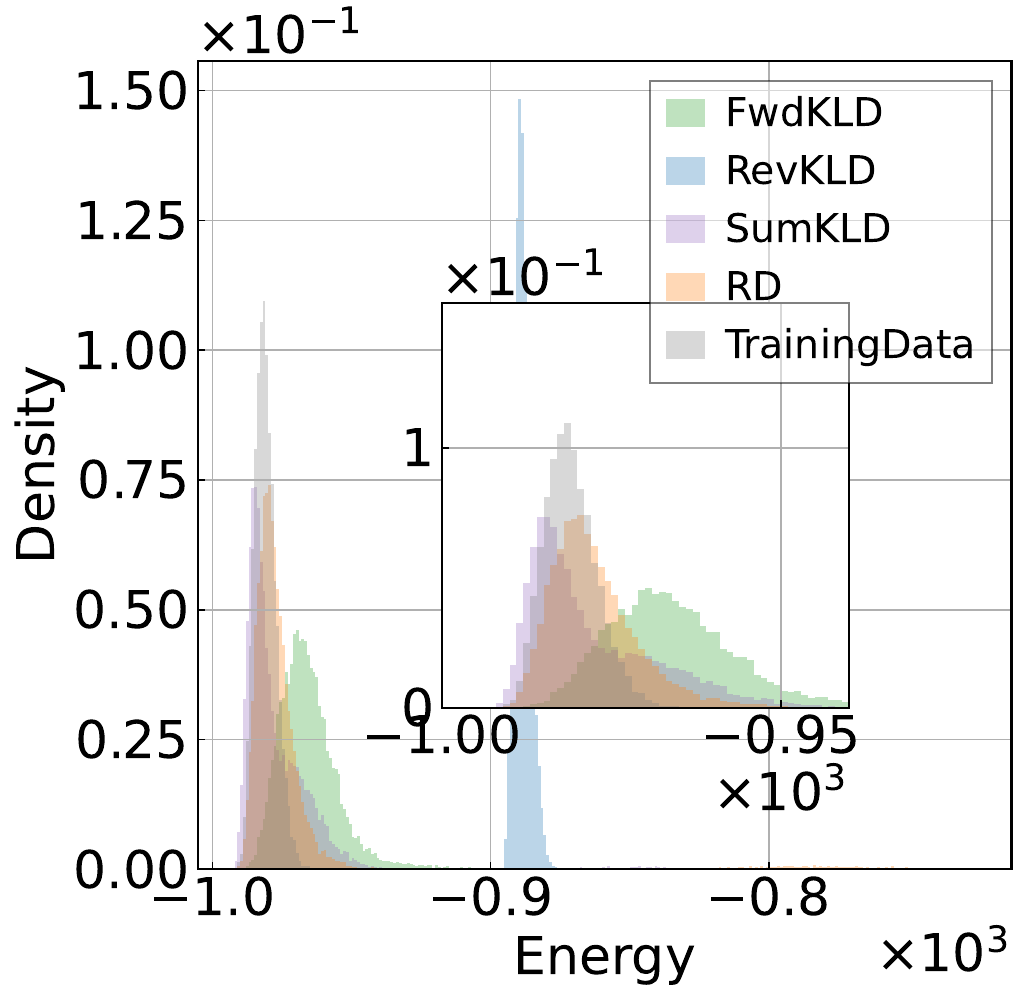}
                    \put(0,95){\subfiglabelsize\captiontext*{}}
                \end{overpic}
            \end{minipage}
        \end{subcaptiongroup}
        \caption{\subref{fig:g18_riw} $R(\theta)$ as functions of epochs during the training process of RBMs and \subref{fig:g18_energy_dist} energy distributions of generated samples by each model and those in the training dataset on Gset G18 of the MCP.}
        \label{fig:g18_riw_and_energy_dist}
    \end{minipage}
    \hfill
    \begin{minipage}[t]{0.49\textwidth}
        \centering
        \begin{subcaptiongroup}
            \begin{minipage}[t]{0.49\textwidth}
                \centering
                \phantomcaption
                \label{fig:g18_pca_fwdkl}
                \begin{overpic}[width=\textwidth]{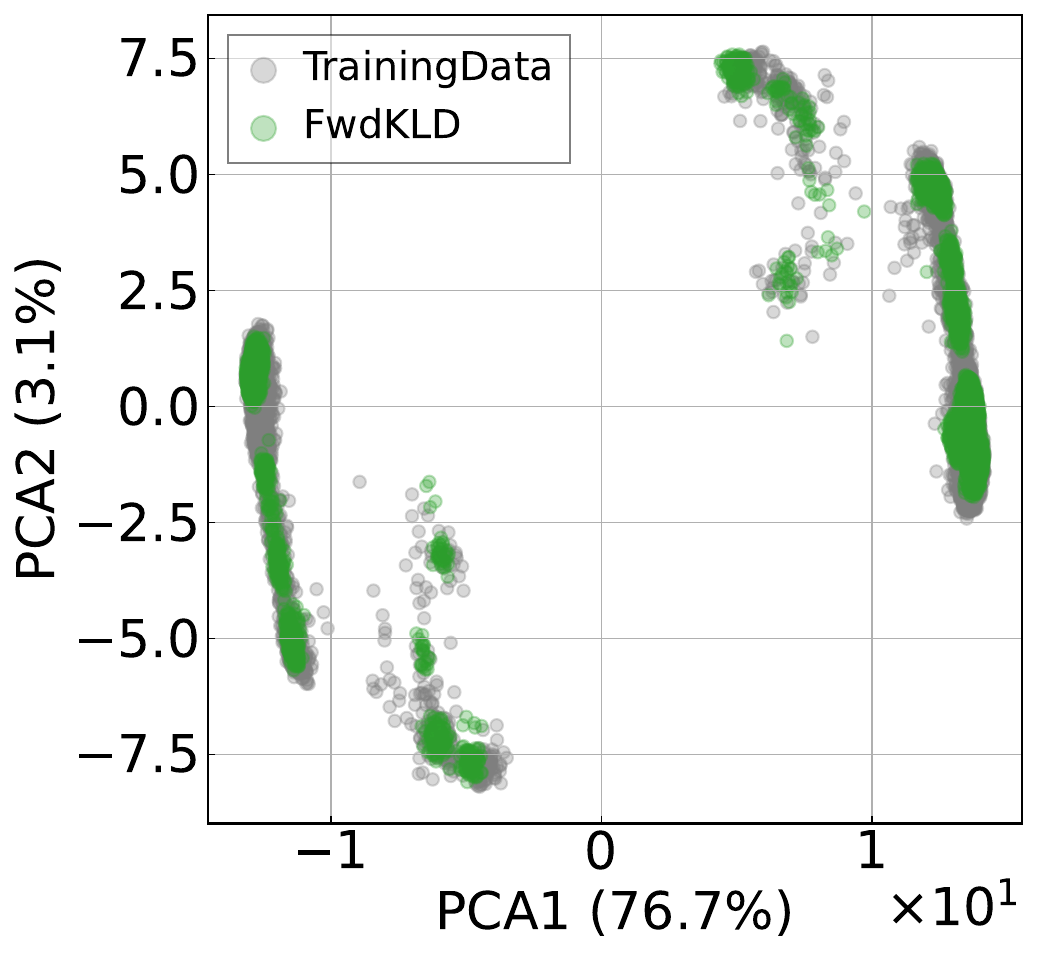}
                    \put(0,95){\subfiglabelsize\captiontext*{}}
                \end{overpic}
            \end{minipage}
            \hfill
            \begin{minipage}[t]{0.49\textwidth}
                \centering
                \phantomcaption
                \label{fig:g18_pca_revkl}
                \begin{overpic}[width=\textwidth]{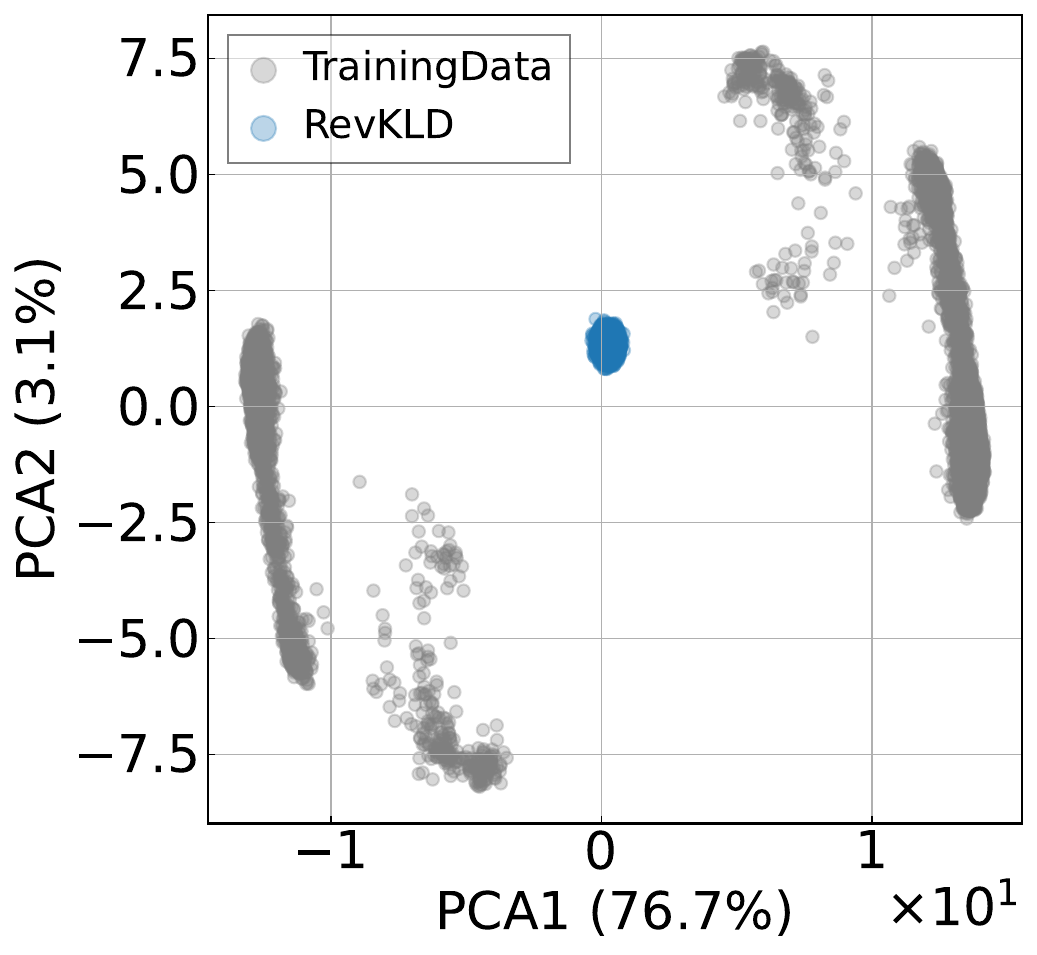}
                    \put(0,95){\subfiglabelsize\captiontext*{}}
                \end{overpic}
            \end{minipage}
            \hfill
            \begin{minipage}[t]{0.49\textwidth}
                \centering
                \phantomcaption
                \label{fig:g18_pca_fwdrevkl}
                \begin{overpic}[width=\textwidth]{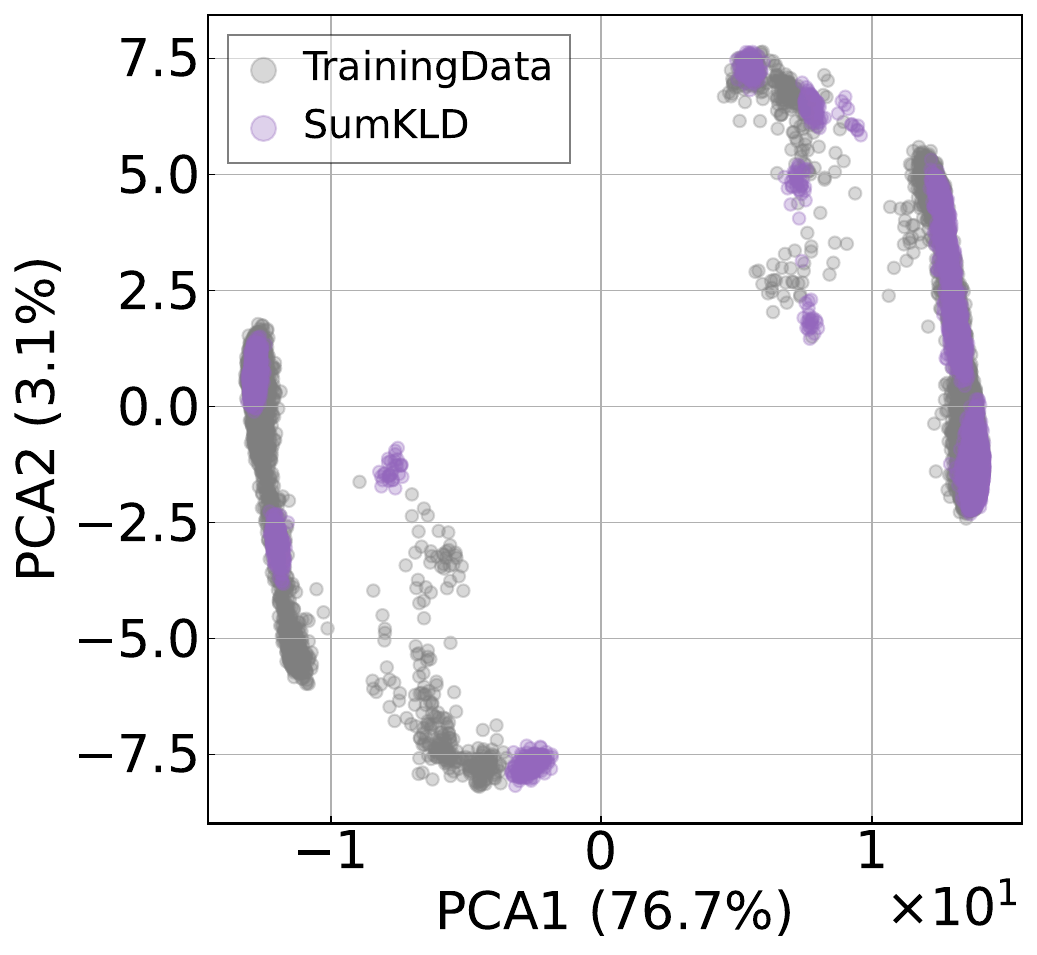}
                    \put(0,95){\subfiglabelsize\captiontext*{}}
                \end{overpic}
            \end{minipage}
            \hfill
            \begin{minipage}[t]{0.49\textwidth}
                \centering
                \phantomcaption
                \label{fig:g18_pca_rd}
                \begin{overpic}[width=\textwidth]{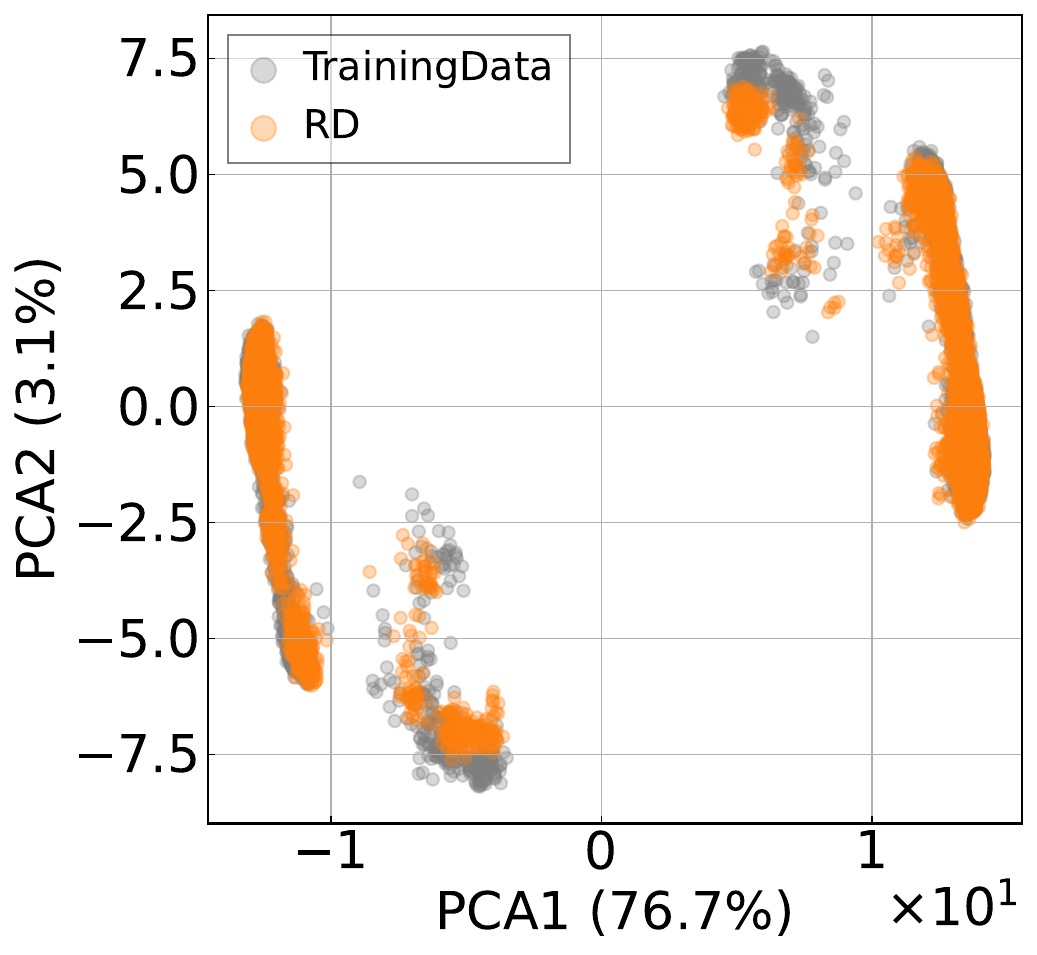}
                    \put(0,95){\subfiglabelsize\captiontext*{}}
                \end{overpic}
            \end{minipage}
        \end{subcaptiongroup}
        \caption{\subref{fig:g18_pca_fwdkl}--\subref{fig:g18_pca_rd} Two dimensional PCA mapping of generated samples by each method for the G18 of MCP.}
        \label{fig:g18_pca}
    \end{minipage}
    \begin{minipage}[t]{0.6\textwidth}
        \centering
        \includegraphics[width=\textwidth]{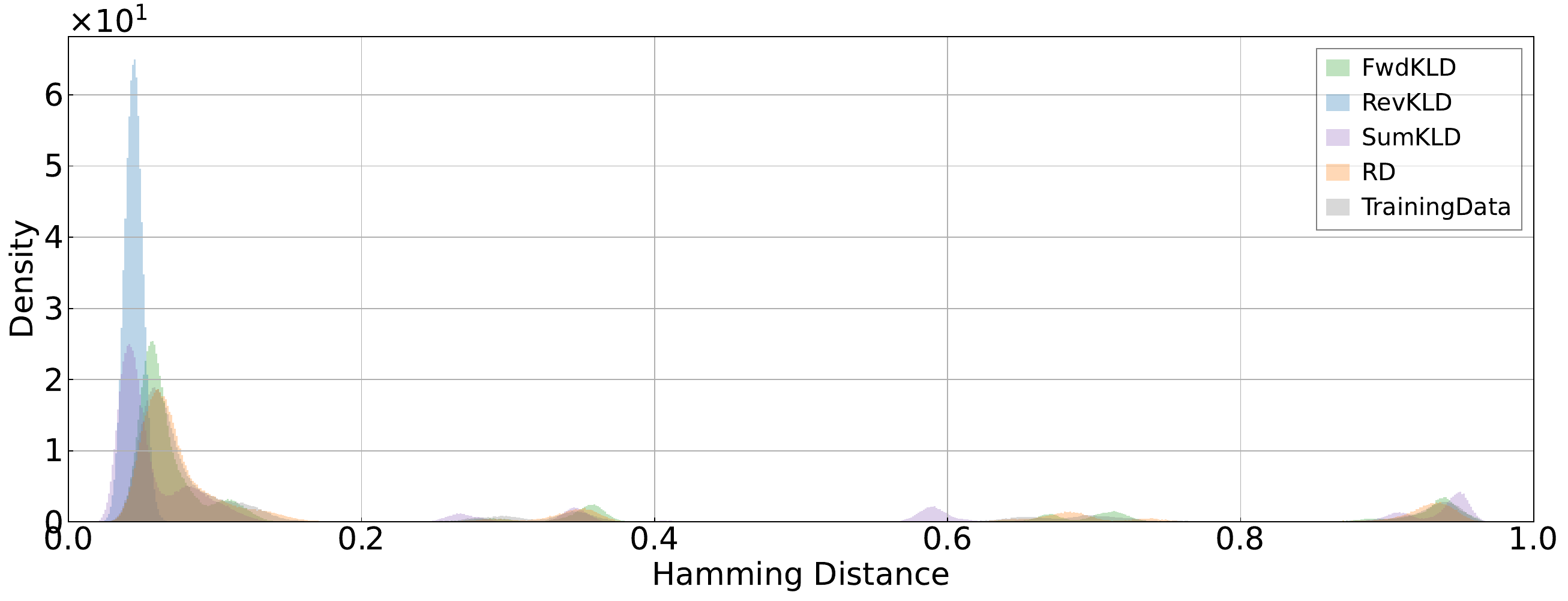}
        \caption{Empirical distribution of the hamming distance of the generated samples by each method and those in the training dataset.}
        \label{fig:g18_hamming_dist}
    \end{minipage}
\end{figure}

As shown in Table~\ref{tab:wasserstein_metric}, the Wasserstein metric of RD learning is not the smallest.
However, as shown in Figure~\ref{fig:g6_riw_and_energy_dist}--\ref{fig:g18_pca}, which is the results of G6, G14, and G18, RD learning significantly outperforms forward KLD learning in terms of $R(\theta)$.
Moreover, the results of projection into reduced feature space obtained by PCA for the training dataset and the hamming distance distribution imply that RD learning is mode-covering.

\end{document}